\documentclass{article}

\usepackage{microtype}
\usepackage{graphicx}
\usepackage{subfigure}
\usepackage{booktabs} 
\usepackage{float}
\usepackage{hyperref}
\usepackage[table,xcdraw]{xcolor}
\usepackage{bm}

\usepackage[accepted]{icml2026}

\usepackage{amsmath}
\usepackage{amssymb}
\usepackage{mathtools}
\usepackage{amsthm}
\usepackage{enumitem}
\setlist{nosep} 
\usepackage[capitalize,noabbrev]{cleveref}

\theoremstyle{plain}

\theoremstyle{definition}

\theoremstyle{remark}

\usepackage{multirow}
\usepackage[textsize=tiny]{todonotes}
\usepackage[normalem]{ulem}
\useunder{\uline}{\ul}{}
\newcommand{\tabref}[1]{Table~\ref{#1}}

\icmltitlerunning{E-VAds: An E-commerce Short Videos Understanding Benchmark for MLLMs}

\begin{document}

\twocolumn[
\icmltitle{E-VAds: An E-commerce Short Videos Understanding Benchmark for MLLMs}



\icmlsetsymbol{equal}{*}
\icmlsetsymbol{intern}{$\dagger$}
\icmlsetsymbol{leader}{$\ddagger$}

\begin{icmlauthorlist}
\icmlauthor{Xianjie Liu}{equal,intern,a}
\icmlauthor{Yiman Hu}{equal,leader,a}
\icmlauthor{Liang Wu}{a}
\icmlauthor{Ping Hu}{c}
\icmlauthor{Yixiong Zou}{b}
\icmlauthor{Jian Xu}{a}
\icmlauthor{Bo Zheng}{a}
\end{icmlauthorlist}

\icmlaffiliation{a}{Alimama Tech, Taobao \& Tmail Group of Alibaba}
\icmlaffiliation{b}{Huazhong University of Science and Technology}
\icmlaffiliation{c}{Vin University}

\icmlcorrespondingauthor{Yiman Hu}{huyiman.hym@alibaba-inc.com}
\icmlcorrespondingauthor{Liang wu}{wuliang.wu@taobao.com.}
\icmlcorrespondingauthor{Jian Xu}{xiyu.xj@taobao.com.}
\icmlcorrespondingauthor{Bo Zheng}{bozheng@alibaba-inc.com}

\icmlkeywords{MLLMs, Benchmark, Video understanding, E-commerce}

\vskip 0.3in
]



\printAffiliationsAndNotice{\icmlEqualContribution, \icmlInternNote, \icmlLeaderNote} 

\begin{abstract}

E-commerce short videos represent a high-revenue segment of the online video industry characterized by a goal-driven format and dense multi-modal signals. Current models often struggle with these videos because existing benchmarks focus primarily on general-purpose tasks and neglect the reasoning of commercial intent. In this work, we first propose a \textbf{multi-modal information density assessment framework} to quantify the complexity of this domain. Our evaluation reveals that e-commerce content exhibits substantially higher density across visual, audio, and textual modalities compared to mainstream datasets, establishing a more challenging frontier for video understanding.
To address this gap, we introduce \textbf{E-commerce Video Ads Benchmark}, which is the first benchmark specifically designed for e-commerce short video understanding. We curated 3,961 high-quality videos from Taobao covering a wide range of product categories and used a multi-agent system to generate 19,785 open-ended Q\&A pairs, which consist of five distinct tasks. Finally, we develop \textbf{E-VAds-R1}, an RL-based reasoning model featuring a multi-grained reward design called \textbf{MG-GRPO}. This strategy provides smooth guidance for early exploration while creating a non-linear incentive for expert-level precision. Experimental results demonstrate that E-VAds-R1 achieves a 109.2\% performance gain in commercial intent reasoning with only a few hundred training samples. Data is available at \url{https://github.com/TaobaoTmall-AlgorithmProducts/E-VAds_Benchmark}.

\end{abstract}

\section{Introduction}
\label{intro}

E-commerce videos have grown rapidly and now represent a major, high-revenue segment of online video. On e-commerce platforms (e.g. TikTok Shop, Amazon, Shopee, Taobao) \cite{ecommerce}, videos are primarily designed to drive immediate purchases rather than general engagement, which makes their style distinct from other video content. In practice, these short ads are brief, fast-paced, and heavily edited, with a clear conversion goal. They pack dense multi-modal signals, such as rapid visual changes, on-screen text, continuous speech, and product close-ups and present them at the same time. This \textbf{multi-modally dense, goal-driven} format introduces new challenges for current video understanding.

Driven by advancements in LLMs, contemporary models excel at general video understanding. However, most efforts remain concentrated on general-purpose \cite{activitynetqa,egoschema,videomme,mvbench,movieqa,nextqa} or long brand advertising \cite{adsqa,videoads} videos, leaving the complex and high-value domain of e-commerce short videos largely under-explored. While existing benchmarks prioritize tasks like action recognition, commonsense QA, grounding and spatial-temporal relationships, they neglect the reasoning of commercial intent, such as selling points, target audiences, conversion strategies and so on.

This domain raises three practical challenges. \textbf{(1) High multi-modal information density:} models must track rapid visual changes while grounding dense speech and text overlays within short time windows. \textbf{(2) Benchmark gap:} there is still no dedicated benchmark that systematically evaluates conversion-oriented e-commerce short videos at scale. \textbf{(3) Open-ended commercial reasoning:} commercial questions (e.g., persuasion logic and consumer insight) are inherently open-ended, highly intent-driven and subjective, making supervision and evaluation less straightforward and often leading to sparse reward signals for learning.

To quantify the above ``high infomation density'' challenge, we further propose a \textbf{multi-modal information density assessment framework} with three complementary metrics: Visual dynamic density ($V_{den}$) \cite{videoads}, which captures the rate of semantic change over time to reflect transition and editing frequencies; Audio density ($A_{den}$), measured by the number of ASR words per unit of time to represent speech intensity; and Textual density ($O_{den}$), defined by the frequency of OCR occurrences per frame to reflect the presence of on-screen text. As shown in \tabref{tab:benchmark_comparison}, \textbf{E-VAds exhibits substantially higher density across vision, audio, and text than mainstream datasets, establishing a more challenging frontier}.

\begin{figure*}
  \centering
  \includegraphics[width=0.9\textwidth]{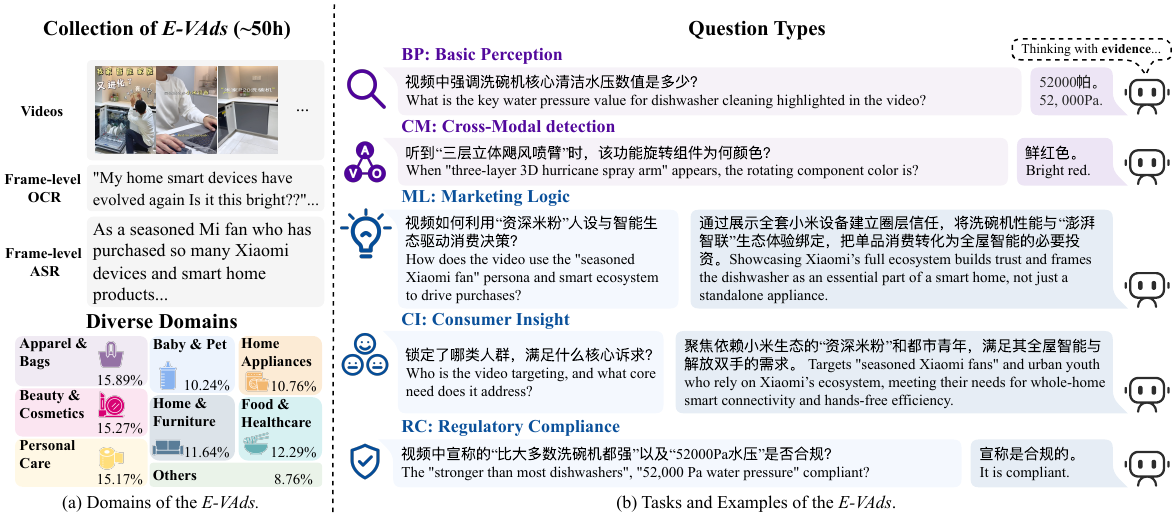}
  \caption{
Overview of \textit{E-VAds} benchmark.
 }
  \label{fig:Overview}
\end{figure*}

To fill this gap, we introduce \textbf{E-commerce Video Ads Benchmark (E-VAds)}, a benchmark for evaluating model performance on e-commerce short video understanding. 
We collect 3{,}961 high-quality videos from Taobao, covering a wide range of product categories, and apply a \textbf{dynamic sampling strategy} to improve category balance and annotation efficiency. Each video is converted into \textbf{a structured multi-modal context} that includes time-aligned ASR and OCR, visual evidence, and metadata. We then generate higher-quality question answering pairs using a \textbf{multi-agent annotation system} to reduce subjectivity in intent-based reasoning tasks for commercial videos. Multi-role agents propose and evaluate candidate QAs, and all items are further verified through rigorous manual review.
The resulting benchmark contains 19{,}785 open-ended question answering pairs across five tasks, spanning two dimensions, \textit{Perception} and \textit{Cognition and Reasoning} (Figure~\ref{fig:Overview}).

\begin{table*}[t!]
\small
    \centering
    \caption{Comparison between E-VAds and other video benchmarks. \textbf{Anno} denotes annotation type  (\textbf{M}: Manual, \textbf{A}: Automatic). \textbf{Task Types} include \textbf{MCQs}  (Multiple-Choice Questions) and \textbf{Open}-ended QA. The best results are in \textbf{bold}, and the second best are \underline{underlined}.}
    \renewcommand{\arraystretch}{0.95}
    \renewcommand{\tabcolsep}{4mm}
    \begin{tabular}{c|ccc|ccc}
        \toprule
\textbf{Benchmark}         & \textbf{QA nums} & \textbf{Anno} & \textbf{Task Types} & $\bm{V_{\mathrm{den}}}$ & $\bm{A_{\mathrm{den}}}$ & $\bm{O_{\mathrm{den}}}$ \\
\hline
VideoMME-short \cite{videomme}            & 3,000            & M             & MCQs                & 31.65                   & 1.16                    & 4.32                    \\
MVBench  \cite{mvbench}                  & 4,000            & M             & MCQs                & 20.69                   & 2.46                    & 3.45                    \\
ActivityNetQA   \cite{activitynetqa}           & {\ul 8,000}            & M             & Open                & 29.69                   & {\ul 2.77}                    & 2.98                    \\
EgoSchema  \cite{egoschema}                & 5,031            & A             & MCQs                & 25.76                   & 0.00                       & 4.24                    \\
AdsQA \cite{adsqa}                     & 7,895            & A+M           & Open                & 50.13                   & 0.85                    & {\ul 5.05}                    \\
VideoAds  \cite{videoads}                 & 1,100            & M             & MCQs                & {\ul 50.72}                   & 1.40                     & 4.02                    \\
\hline
\rowcolor{gray!20}
\textbf{E-VAds (Ours)} & \textbf{19,785}           & A+M           & Open                & \textbf{60.44}                   & \textbf{5.08}                    & \textbf{18.66}                   \\
\hline
\textit{E-VAds-Test} (Ours)         & 16,925           & A+M           & Open                & 60.48                   & 5.09                    & 18.81                   \\
\textit{E-VAds-RL} (Ours)           & 980              & A+M           & Open                & 59.92                   & 5.17                    & 19.88                   \\
\textit{E-VAds-SFT} (Ours)          & 1,880            & A+M           & Open                & 60.40                    & 5.00                       & 16.60            \\
        \bottomrule
    \end{tabular}
    \label{tab:benchmark_comparison}
\end{table*}

Finally, we propose \textbf{E-VAds-R1}, an RL-based reasoning model to handle the modality-dense videos and the complex open-ended commercial questions. We design evidence-grounded rewards that encourage multi-modal attribution, and introduce \textbf{MG-GRPO}, a multi-grained reward design that ensembles reward granularities to provide \textbf{smooth guidance} during early exploration while creating a \textbf{non-linear incentive} for expert-level precision. With only a few hundred training samples, E-VAds-R1 achieves a significant 109.2\% relative improvement in commercial intent reasoning over strong general-purpose baselines.

Our main contributions are as follows:
\begin{itemize}[noitemsep, topsep=2pt, leftmargin=*]
    \item We introduce \textbf{E-VAds}, the first benchmark for e-commerce short video understanding, with an automated construction pipeline to complement existing video benchmarks.
    \item We propose a \textbf{multi-modal information density assessment framework} and show that E-VAds contains much denser visual, audio, and textual information than mainstream datasets, making it more challenging for MLLMs' understanding.
    \item We develop \textbf{E-VAds-R1}, an RL-based reasoning model with a multi-grained reward design, achieving a 109.2\% performance gain in the e-commerce domain.
\end{itemize}

\section{Related Works}
\label{related}

\subsection{Video Question Answering Benchmarks}
VideoQA benchmarks evaluate spatiotemporal understanding in videos. Representative datasets include human-centered benchmarks such as NextQA \cite{nextqa} and MovieQA \cite{movieqa}, as well as instructional benchmarks such as EgoSchema \cite{egoschema} and VideoMME \cite{videomme}. However, these general benchmarks rarely capture the persuasive logic and conversion-oriented mechanisms central to advertising. While AdsQA \cite{adsqa} and VideoAds \cite{videoads} study advertising videos, they mainly focus on longer, carefully produced brand ads for brand awareness building, and largely overlook e-commerce short videos that target immediate conversion and contain tightly synchronized, multimodally dense signals.

\subsection{Video Large Language Models}
MLLMs build on vision language alignment from CLIP \cite{clip}, with models such as LLaVA \cite{llava} and Flamingo \cite{flamingo} connecting vision encoders to LLMs for instruction following. This line has extended to video via Video-LLaVA \cite{videollava}, vidoe-LLama \cite{videollamav3} and VideoChat \cite{videochat} using temporal aggregation for multi-frame modeling. Closed-source models including the GPT series \cite{GPT52} and Gemini \cite{Gemini3} further improve general multimodal reasoning, while InternVL \cite{internvl3,internvl35} and QwenVL \cite{qwen25vl,qwen3vl} enhance fine-grained perception and long-context reasoning \cite{feng2026interactive}. Recently, some models have also shown competitive performance, such as Keye \cite{keye-vl,keye15} and Mimo \cite{mimo}. Despite this progress, existing open-source models still struggle with the high-density multimodal signals common in e-commerce videos.

\subsection{Reinforcement Learning for Reasoning}
In text-only settings, supervised fine-tuning is often constrained by data diversity, whereas RLHF \cite{rlhf} better aligns outputs with human preferences. Recent results from DeepSeek-R1 \cite{GRPO} and OpenAI o1 \cite{openai-o1} suggest that reinforcement signals can strengthen reasoning. In advertising, AdsQA \cite{adsqa} proposes ReAd-R, using an LLM-as-a-judge \cite{llmasjudge} to guide reflection on social intent. However, it mainly targets brand ads and focuses on metaphor and emotional tone, and does not emphasize the multimodal evidence needed for commercial intent reasoning in dense e-commerce videos such as those in E-VAds.

\section{Multi-modal Information Density Assessment Framework}
\subsection{Definition of Multi-modal Information Density}
To quantify information density and multi-modal complexity in e-commerce short videos, we define three modality-specific metrics that capture visual dynamics, spoken content, and on-screen text.

\paragraph{Visual dynamic density ($V_{\mathrm{den}}$).}
Following \cite{videoads}, we use DINOv3-Base to extract frame features $f$. For a video with $T$ sampled frames, we compute the weighted average similarity of frame $i$ within a temporal neighborhood of size $d$:
\begin{equation}
\bar{S}_i=\frac{\sum_{j \in N_i, j \neq i} w(i,j)\cdot \cos (f_i,f_j)}{\sum_{j \in N_i, j \neq i} w(i,j)} ,
\end{equation}
where $\cos(\cdot)$ denotes cosine similarity and $w(i,j)=\exp\!\left(-\frac{|j-i|}{2d}\right)$ applies exponential temporal decay. We then define visual dynamic density as
\begin{equation}
V_{\mathrm{den}}=\alpha \cdot \frac{1}{T}\sum_{i=1}^{T}\left(1-\bar{S}_i\right),
\end{equation}
where $\alpha$ is a scaling constant set to 100. A larger $V_{\mathrm{den}}$ indicates more frequent visual changes and stronger editing dynamics.

\begin{figure*}[t!]
  \centering
  \includegraphics[width=0.9\textwidth]{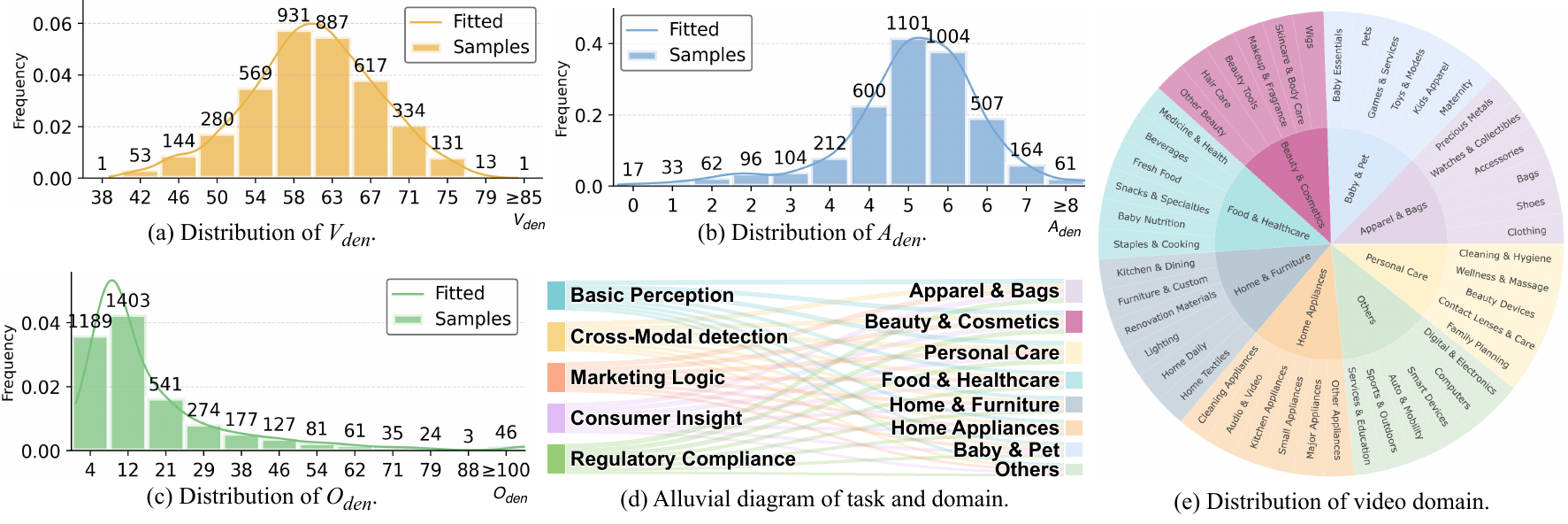}
  \caption{
Statistics of E-VAds benchmark.
 }
  \label{fig:domain}
\end{figure*}
\paragraph{Audio density ($A_{\mathrm{den}}$) and Textual density ($O_{\mathrm{den}}$).}
We define audio and textual density as the word counts of ASR and OCR content normalized by video duration:
\begin{equation}
A_{\mathrm{den}}=\frac{\lvert \mathcal{T}_{\mathrm{asr}} \rvert}{T}, \qquad
O_{\mathrm{den}}=\frac{\lvert \mathcal{T}_{\mathrm{ocr}} \rvert}{T},
\end{equation}
where $\mathcal{T}_{\mathrm{asr}}$ is the full-video ASR transcript, $\mathcal{T}_{\mathrm{ocr}}$ is the concatenated OCR text from sampled frames, $\lvert\cdot\rvert$ denotes word count, and $T$ is the video duration (seconds). Larger $A_{\mathrm{den}}$ and $O_{\mathrm{den}}$ indicate denser speech and on-screen text, respectively.


Together, these metrics provide a unified view of how e-commerce ads present dense information across vision, audio, and text compared with general videos.

\subsection{Metric Analysis and Comparison}
\tabref{tab:benchmark_comparison} compares E-VAds with representative video understanding and advertising benchmarks in terms of scale, task format, and multi-modal information complexity. Across all three dimensions, E-VAds exhibits substantially higher density than both general video QA datasets and existing advertising benchmarks. Figure~\ref{fig:distribution} and Appendix.~\ref{app:distribution} show the detailed distributions of the three density metrics in E-VAds.

These results confirm that e-commerce short videos are not a minor variant of general video QA, but \textbf{a qualitatively harder setting where models must operate under modality-dense and tightly synchronized signals.} In practice, models must \emph{simultaneously} (i) track rapid visual changes, (ii) associate fast-evolving ASR/OCR cues with the correct visual evidence, and (iii) reason about commercial intent when signals are noisy, partially redundant, or even conflicting. Moreover, the metrics provide a principled way to quantify this difficulty and enable density-aware analyses, which were not provided in prior datasets.

\section{The E-VAds Benchmark}
To fill the gap in current video benchmarks, we introduce \textbf{E-VAds}, a benchmark designed to evaluate MLLMs' commercial understanding in conversion-oriented e-commerce short videos, which is an important and challenging domain.

In Section~3.1, we present fine-grained modality extraction, decomposition, and alignment to better handle high-density multi-modal signals. In Sections~3.2 and~3.3, to improve the reliability of the generated open-ended commercial-intent QA pairs, we adopt a multi-agent system for multi-round role-based generation and human expert review, and construct multi-modal evidence chain \cite{s1,deepseek} to support reasoning. The benchmark construction pipeline is shown in Figure~\ref{fig:Pipeline} and the statistics are shown in Figure~\ref{fig:distribution}.

\subsection{Data Collection and Multi-Modal Alignment}
\paragraph{Data collection and filtering.}
To select high-quality data from billions of Taobao e-commerce advertising videos while covering all video categories as much as possible, we design an automated filtering pipeline that removes videos with weak commercial appeal, low-quality or overly short videos and samples with missing metadata, resulting in about 30{,}000 high-quality promotional videos.

\begin{figure}[t!]
  \centering
  \includegraphics[width=0.48\textwidth]{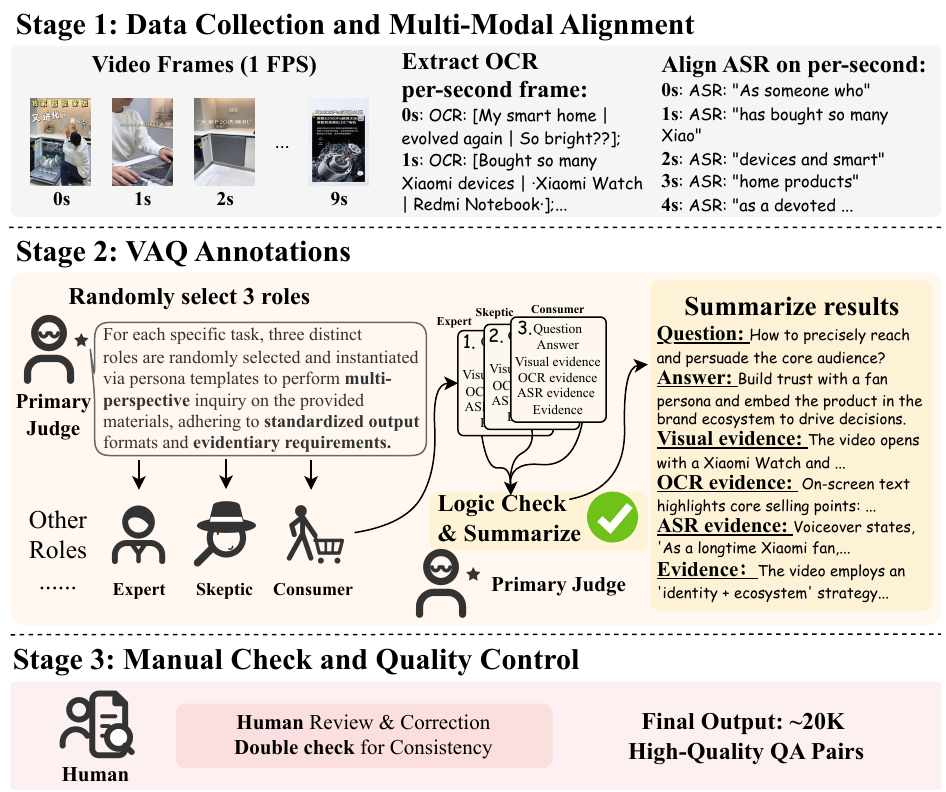}
  \caption{
Dataset Construction Pipeline.
 }
  \label{fig:Pipeline}
\end{figure}

We additionally propose a \textbf{dynamic sampling algorithm} to reduce annotation cost and alleviate category imbalance. Instead of uniform undersampling, we use a sigmoid-based function to set the sampling ratio $f(x)$ for each category:
\begin{equation}
f(x) = \frac{a}{1 + \exp\left(1-\frac{b}{x}\right)},
\end{equation}
where $x$ is the original number of videos in a category, $a$ is the upper bound of the sampling ratio, and $b$ controls the curve’s inflection point. \textit{This function preserves minority categories with higher sampling ratios while suppressing majority categories non-linearly}. After sampling, we obtain 3{,}961 videos with a more balanced category distribution (Fig.~\ref{fig:domain} (d)) and improved annotation efficiency.

\paragraph{Dense multi-modal signal extraction, decomposition and alignment.}
E-commerce short videos are produced to drive immediate conversions by coordinating persuasive cues across visual, audio, and textual modalities. To better leverage such carefully crafted videos for fine-grained analysis, we slice each raw video into \textbf{a compact, time-aligned event sequence} (Figure~\ref{fig:Pipeline}) that is easy for both annotators and models to inspect. We sample frames at 1 FPS, transcribe speech with Whisper-v3-large \cite{whisperv3}, and extract on-screen text with Qwen2.5-VL 32B \cite{qwen25vl}. We align all modalities on a 1-second timeline: OCR is assigned to its corresponding second, while each ASR segment is split into per-second non-overlapping chunks according to its temporal span (the last second absorbs remaining characters to avoid truncation). We then merge consecutive seconds with identical OCR and ASR content to reduce redundancy.

We formalize the structured context as
\begin{equation}
C = \{\langle f_t, \alpha_t, \gamma_t \rangle \mid t = 1, \dots, T\} \otimes \mathcal{M},
\end{equation}
where $f_t$ is the visual keyframe feature at time $t$, $\alpha_t$ is the aligned speech text within $[t, t+1]$, $\gamma_t$ is the calibrated and de-duplicated OCR text, $\mathcal{M}$ is product metadata such as category and attributes, and $\otimes$ denotes temporal alignment and semantic concatenation across modalities. \textbf{This pipeline converts noisy multimodal streams into a structured evidence chain for downstream persuasive-content analysis.}

\subsection{VQA Annotations}
\paragraph{Task definition.}
\label{par: task definition}
E-commerce videos pack dense product details and promotional claims into a few seconds through coordinated visuals, narration, and on-screen text. As a result, models must first accurately recognize fine-grained concepts, and then reason about intent, audience, and compliance based on multi-modal evidence. Therefore, we design our tasks along two dimensions, perception and reasoning, to match the characteristics of e-commerce short videos. For perception, in addition to basic tasks that assess core recognition abilities, we introduce a cross-modal detection task to evaluate how well models handle the high-density multi-modal signals across modalities. For reasoning, we design three groups of tasks from the perspectives of advertisers, consumers, and platforms. In total, we define five task categories, summarized below, with detailed prompts provided in Appendix~\ref{appendix:task_prompts}.


Dimension 1: Perception.
\begin{itemize}[noitemsep, topsep=0pt, leftmargin=*]
    \item Basic Perception (BP): Identify product attributes and salient visual entities.
    \item Cross-Modal Detection (CM): Judge consistency and complementarity among ASR, OCR, and visual cues under noise.
\end{itemize}

Dimension 2: Cognition and Reasoning.
\begin{itemize}[noitemsep, topsep=0pt, leftmargin=*]
    \item Marketing Logic (ML): Unpack persuasive structure, including selling points and pain point to solution mapping.
    \item Consumer Insight (CI): Infer target audience from style, tone, and product characteristics.
    \item Regulatory Compliance (RC): Identify potential violations of advertising regulations.
\end{itemize}

\paragraph{Multi-agent annotation system.}
Due to the nature of e-commerce tasks, they are often open-ended and involve intent-driven subjective Q\&As. To generate fairer and more objective Q\&A pairs for such tasks, we design a multi-agent annotation system \cite{evoagent} that arbitrates among responses from diverse roles and enforces supervision with \textbf{multi-modal evidence} refined from the time-aligned multi-modal sequence. 
During the annotation, each agent adopts a distinct commercial persona and proposes challenging, evidence-grounded questions based on the structured context $C$, while a primary judge moderates the process with support from multiple secondary roles.
Detailed definitions are provided in Appendix~\ref{appendix:multiagent}.


\textbf{Traceability constraint.}
To ensure reproducibility and prevent impression-based answers, each QA must satisfy a strict traceability rule:
\[
\texttt{Evidence(V, A, O)} \rightarrow \texttt{Reasoning} \rightarrow \texttt{Answer}.
\]
Each answer must be supported by at least one evidence source: vision (V), ASR (A), or OCR (O).

\textbf{Cross-modal detection constraint.}
For CM, we further enforce an \emph{information-gap} design: one modality raises the question and another modality provides decisive evidence.
This discourages single-modality shortcuts and explicitly tests cross-modal retrieval and alignment \cite{feng2026robust}.

\textbf{Question normalization.}
We cap question length, remove leading phrasing and hints, and rewrite questions into concise and objective academic language so that performance depends on evidence-based inference.




\subsection{Manual Check and Quality Control}
After the automated annotation process, we implemented a review mechanism combining manual selection and a cycle-based elimination mechanism to ensure factual accuracy, clarity of expression, and consistency in difficulty. Appendix~\ref{appendix:manual_check} shows the details of the review process, including the annotation interface, detailed reviewer checklists, and complete manual verification specifications.
After manual check and quality control, E-VAds ultimately contains 19,785 high-quality QA pairs from 3,961 videos.

\section{The E-VAds-R1 Model}
Motivated by the observed density-induced challenges and the open-ended nature of commercial reasoning, our subsequent E-VAds-R1 study further explores how to improve learning under sparse supervision by designing multi-grained rewards for reinforced fine-tuning.

\begin{figure}[t!]
  \centering
  \includegraphics[width=0.48\textwidth]{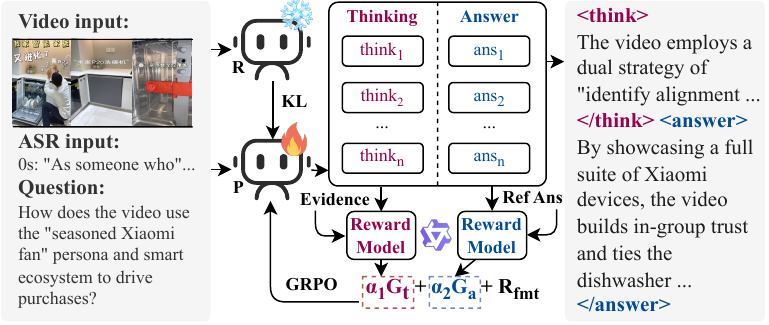}
  \caption{
In the E-VAds-R1 framework, given a question, the policy model produces multiple responses including think and answer; these are scored by a reward model, and the resulting rewards guide policy updates through policy gradient optimization.
 }
  \label{fig:training}
\end{figure}

Figure~\ref{fig:training} shows the training framework of E-VAds-R1. Given a video and a question, the policy outputs \texttt{<think>} and \texttt{<answer>}, and a frozen LLM-as-a-judge provides a scalar reward to optimize the policy.

\subsection{Data Splits and Output Format}
The training set is split into E-VAds-Train-SFT (376 videos, 1{,}980 QA) for supervised instruction and format alignment, and E-VAds-Train-RL (196 videos, 980 QA) for reinforcement learning in complex commercial scenarios. The remaining 3{,}389 videos with 16{,}384 QA pairs form the E-VAds test split. All training samples follow a structured format:
\texttt{<think>} and \texttt{<answer>}.

\subsection{Training Pipeline}
To better handle the complex commercial tasks in E-VAds, we use a two-stage pipeline from imitation to reinforcement learning.
In the SFT stage, we convert E-VAds annotations into instruction-style samples that require explicit evidence grounding before answering, aligning the model with e-commerce semantics and enforcing the structured output.
The resulting model learns basic e-commerce video understanding and cross-modal grounding.
We then apply RL to improve attribution and reasoning consistency by rewarding outputs that are evidence-grounded, logically coherent, and explicitly link visual, ASR, and OCR cues to commercial intent.
The prompting details are provided in Appendix~\ref{appendix:sft_rl_prompts}.

\subsection{Reward Design}
\label{sec:RL}
We use an LLM-as-a-judge \cite{llmasjudge,funqa,canllmasjudge,autoeval} to compare model predictions against expert-annotated ground truth (GT) and output a five-level score: $x \in \{0, 0.25, 0.5, 0.75, 1\}$. For each response, the judge verifies the answer against evidence from all three modalities: visual, ASR, and OCR, as well as the final evidence summary. We report three metrics:
\begin{itemize}[noitemsep, topsep=2pt, leftmargin=*]
    \item \textbf{Strict ($S$):} $S(x)=\mathbb{I}(x=1)$.
    \item \textbf{Relaxed-3 ($R3$):} $R3(x)=1$ if $x=1$; $R3(x)=0.5$ if $x\in\{0.75,0.5\}$; otherwise $R3(x)=0$.
    \item \textbf{Relaxed-5 ($R5$):} $R5(x)=x$.
\end{itemize}

Here $\mathbb{I}(\cdot)$ is the indicator function. All metrics are averaged over samples. We provide the judge prompt, rubric, and examples in Appendix~\ref{appendix:judge_prompt}.

During reinforcement learning, we score each generated trace along the following dimensions:
\begin{enumerate}[noitemsep, topsep=2pt, leftmargin=*]
    \item Reasoning trace: quality of the thinking ($x_t$).
    \item Terminal answer: quality of the answer ($x_a$).
    \item Format constraint ($R_{\mathrm{fmt}}$): $R_{\mathrm{fmt}}=-1$ if required tags are missing or malformed, otherwise $0$.
\end{enumerate}

\begin{table*}[t!]
\centering
\caption{Benchmark results for different MLLMs. * means 400 randomly sampled questions.}
\renewcommand{\arraystretch}{0.95}
\renewcommand{\tabcolsep}{0.12mm}
\begin{tabular}{c|c|cccccc|cccccc|cccccc}
\toprule
                                         &                          & \multicolumn{6}{c|}{\textbf{S}}                                                                                                & \multicolumn{6}{c|}{\textbf{R3}}                                                                                               & \multicolumn{6}{c}{\textbf{R5}}                                                                                               \\
\multirow{-2}{*}{\textbf{Model}}      & \multirow{-2}{*}{\textbf{Params}} & \textbf{BP}            & \textbf{CM}            & \textbf{ML}            & \textbf{CI}            & \textbf{RC}            & {\color[HTML]{3076B6} \textbf{ALL}}           & \textbf{BP}            & \textbf{CM}            & \textbf{ML}            & \textbf{CI}            & \textbf{RC}            & {\color[HTML]{3076B6} \textbf{ALL}}           & \textbf{BP}            & \textbf{CM}            & \textbf{ML}            & \textbf{CI}            & \textbf{RC}            & {\color[HTML]{3076B6} \textbf{ALL}}           \\
\hline
Expert$^*$ &  - & .786 & .536 & .173 & .250 & .804 & {\color[HTML]{3076B6} .535} & .893 & .768 & .587 & .625 & .902 & {\color[HTML]{3076B6} .767} & .932 & .839 & .793 & .813 & .951 & {\color[HTML]{3076B6} .871}
\\
\hline
\rowcolor[HTML]{EFEFEF} 
\multicolumn{20}{c}{\textbf{Closed-source Models}}                                                                                                                                                                                                                                                                                                                                                                                                                   \\
\hline
GPT 5.2                                           & -                        & .687          & .315          & .205          & .074          & .105          & {\color[HTML]{3076B6} .278}          & .773          & .526          & .596          & .528          & .268          & {\color[HTML]{3076B6} .539}          & .842          & .671          & .786          & .746          & .477          & {\color[HTML]{3076B6} .704}          \\
Gemini3-Flash                                      & -                        & .763          & .391          & .179          & .193          & .222          & {\color[HTML]{3076B6} .350}          & .836          & .602          & .583          & .588          & .452          & {\color[HTML]{3076B6} .612}          & .888          & .730          & .776          & .776          & .634          & {\color[HTML]{3076B6} .761}          \\
\hline
\rowcolor[HTML]{EFEFEF} 
\multicolumn{20}{c}{\textbf{Open-source Models}}                                                                                                                                                                                                                                                                                                                                                                                                                      \\
\hline
Qwen3-Omni-Instruct                   & \multirow{1}{*}{30BA3B} & .693          & .227          & .034          & .032          & .037          & {\color[HTML]{3076B6} .205}          & .801          & .454          & .508          & .502           & .306          & {\color[HTML]{3076B6} .514}          & .864          & .596          & .701          & .692          & .512          & {\color[HTML]{3076B6} .673}          \\
Qwen3-Omni-Thinking                   & 30BA3B & .637          & .214          & .065          & .054          & .065          & {\color[HTML]{3076B6} .207}          & .755          & .379          & .525          & .517           & .315          & {\color[HTML]{3076B6} .498}          & .835          & .552          & .739          & .721          & .528          & {\color[HTML]{3076B6} .675}          \\
VideoLlama3                           & 7B     & .306          & .082          & .000          & .000          & .000          & {\color[HTML]{3076B6} .078}          & .448          & .208          & .259          & .286           & .049          & {\color[HTML]{3076B6} .250}          & .578          & .382          & .397          & .404          & .232          & {\color[HTML]{3076B6} .399}          \\
InternVL3                             & 8B     & .531          & .093          & .000          & .001          & .002          & {\color[HTML]{3076B6} .126}          & .674          & .266          & .396          & .410           & .153          & {\color[HTML]{3076B6} .380}          & .757          & .420          & .478          & .478          & .343          & {\color[HTML]{3076B6} .495}          \\
InternVL3.5                           & 8B     & .522          & .084          & .004          & .005          & .003          & {\color[HTML]{3076B6} .124}          & .664          & .268          & .483          & .480           & .146          & {\color[HTML]{3076B6} .408}          & .759          & .423          & .629          & .621          & .347          & {\color[HTML]{3076B6} .556}          \\
Keye-VL                               & 8B     & .307          & .080          & .068          & .038          & .075          & {\color[HTML]{3076B6} .119}          & .408          & .509          & .182          & .243           & .495          & {\color[HTML]{3076B6} .373}          & .568          & .712          & .386          & .462          & .707          & {\color[HTML]{3076B6} .572}          \\
Keye1.5-VL                            & 8B     & .619          & .125          & .193          & .051          & .039          & {\color[HTML]{3076B6} .206}          & .726          & .552          & .377          & .245           & .506          & {\color[HTML]{3076B6} .482}          & .805          & .756          & .532          & .463          & .695          & {\color[HTML]{3076B6} .650}          \\
Mimo-VL-SFT                           & \multirow{1}{*}{8B}     & .519          & .007          & .124          & .008          & .006          & {\color[HTML]{3076B6} .133}          & .664          & .486          & .300          & .247           & .479          & {\color[HTML]{3076B6} .435}          & .763          & .676          & .471          & .442          & .616          & {\color[HTML]{3076B6} .594}          \\
Mimo-VL-RL                            & 8B     & .600          & .146          & .009          & .007          & .010          & {\color[HTML]{3076B6} .155}          & .734          & .349          & .491          & .490           & .258          & {\color[HTML]{3076B6} .464}          & .814          & .514          & .698          & .648          & .463          & {\color[HTML]{3076B6} .628}          \\
                     {Qwen2.5-VL-Instruct}  & 32B    & .550          & .150          & .001          & .003          & .003          & {\color[HTML]{3076B6} .142}          & .704          & .354          & .485          & .490           & .289          & {\color[HTML]{3076B6} .464}          & .799          & .523          & .695          & .680          & .478          & {\color[HTML]{3076B6} .635}          \\
\multirow{-1}{*}{Qwen2.5-VL-Instruct} & 7B     & .446          & .108          & .000          & .000          & .001          & {\color[HTML]{3076B6} .111}          & .608          & .268          & .374          & .390           & .206          & {\color[HTML]{3076B6} .369}          & .714          & .437          & .458          & .468          & .368          & {\color[HTML]{3076B6} .489}          \\
\hline
\textbf{E-VAds-R1(Qwen2.5-VL)}                             & \textbf{7B}     & \textbf{.569} & \textbf{.127} & \textbf{.069} & \textbf{.021} & \textbf{.177} & {\color[HTML]{3076B6} \textbf{.193}} & \textbf{.714} & \textbf{.315} & \textbf{.523} & \textbf{.496}  & \textbf{.457} & {\color[HTML]{3076B6} \textbf{.501}} & \textbf{.804} & \textbf{.497} & \textbf{.739} & \textbf{.711} & \textbf{.652} & {\color[HTML]{3076B6} \textbf{.680}} \\
\hline
                Qwen3-VL-Instruct   & 30BA3B & .672          & .236          & .023          & .022          & .027          & {\color[HTML]{3076B6} .196}          & .783          & .467          & .501          & .496           & .253          & {\color[HTML]{3076B6} .500}          & .852          & .609          & .689          & .680          & .472          & {\color[HTML]{3076B6} .660}          \\
\multirow{-1}{*}{Qwen3-VL-Instruct}   & 8B     & .600          & .133          & .006          & .010          & .017          & {\color[HTML]{3076B6} .153}          & .745          & .365          & .490          & .490           & .329          & {\color[HTML]{3076B6} .484}          & .828          & .512          & .623          & .644          & .511          & {\color[HTML]{3076B6} .623}          \\
                 Qwen3-VL-Thinking   & 30BA3B & .566          & .144          & .011          & .007          & .012          & {\color[HTML]{3076B6} .148}          & .711          & .342          & .497          & .490           & .262          & {\color[HTML]{3076B6} .460}          & .803          & .517          & .717          & .661          & .473          & {\color[HTML]{3076B6} .634}          \\
\multirow{-1}{*}{Qwen3-VL-Thinking}   & 8B     & .605          & .172          & .051          & .051          & .051          & {\color[HTML]{3076B6} .186}          & .723          & .349          & .519          & .514           & .307          & {\color[HTML]{3076B6} .482}          & .810          & .523          & .738          & .717          & .514          & {\color[HTML]{3076B6} .660}          \\
\hline
\textbf{E-VAds-R1(Qwen3-VL)}                        & \textbf{8B}     & \textbf{.628} & \textbf{.252} & \textbf{.313} & \textbf{.126} & \textbf{.279} & {\color[HTML]{3076B6} \textbf{.320}} & \textbf{.741} & \textbf{.427} & \textbf{.651} & \textbf{..555} & \textbf{.515} & {\color[HTML]{3076B6} \textbf{.577}} & \textbf{.824} & \textbf{.592} & \textbf{.816} & \textbf{.761} & \textbf{.689} & {\color[HTML]{3076B6} \textbf{.736}}
\\  
\bottomrule
\end{tabular}
\label{tab:compare}
\end{table*}

\paragraph{MG-GRPO.}
To mitigate sparse rewards in open-ended commercial reasoning, we propose \textbf{Multi-Grained GRPO (MG-GRPO)}. It extends GRPO \cite{GRPO} by introducing a multi-grained reward mapping that calibrates responses across different levels of strictness:
\begin{equation}
G(x)=\frac{1}{3}\bigl[S(x)+R3(x)+R5(x)\bigr].
\end{equation}
By combining strict and relaxed scoring, $G(x)$ provides informative rewards for partially correct traces while still strongly favoring fully correct and well-grounded answers. 
\begin{itemize}[noitemsep, topsep=2pt, leftmargin=*]
    \item Smooth Guidance for Exploration: By incorporating the relaxed metric $R5(x)$, $G(x)$ provides dense feedback for partially correct traces. For instance, a marginal improvement from $x=0$ to $x=0.25$ yields a non-zero reward ($G(x) \approx 0.083$), preventing the policy from being lost in a ``zero-reward landscape'' during early stages.
    \item Non-linear Incentive for Precision: The mapping creates an uneven reward landscape to penalize ``near-misses.'' While the reward increment from $x=0.5$ to $x=0.75$ is relatively small ($\approx 0.084$), the leap from $x=0.75$ ($G(x) \approx 0.417$) to a perfect trace ($x=1, G(x)=1.0$) is significantly larger. This non-linear jump, amplified by the strict metric $S(x)$, compels the model to pursue expert-level precision rather than settling for partially grounded reasoning.
\end{itemize}
The final reward $R$ is a weighted combination of the answer score ($x_a$), the reasoning trace score ($x_t$) and a format constraint penalty:
\begin{equation}
R=\alpha_1\,G(x_a)+\alpha_2\,G(x_t)+R_{\mathrm{fmt}},
\end{equation}
where $\alpha_1=0.8$ and $\alpha_2=0.2$. Following GRPO \cite{GRPO}, we sample $n$ traces $\{o_1,\dots,o_n\}$ per prompt and compute the group-normalized advantage:
\begin{equation}
A_i=\frac{R_i-\mathrm{mean}(\{R_1,\dots,R_n\})}{\mathrm{std}(\{R_1,\dots,R_n\})+\epsilon}.
\end{equation}
This multi-grained reward structure enhances group-relative discriminability: by providing a denser reward spectrum, it enables $A_i$ to capture subtle quality differences among traces within the same group, even when none are perfectly correct. This stabilizes the optimization and jointly emphasizes better reasoning paths and terminal answers especially for e-commerce video tasks.

\section{Experiment}
\subsection{Experimental settings}
To ensure reproducibility and fairness, we choose Qwen2.5-VL 7B Instruct and Qwen3-VL 8B Instruct as base models. Training uses 16 H20 GPUs with Llama-Factory (SFT) \cite{llamafactory} and ROLL (RL) \cite{ROLL} Framework. SFT uses a batch size of 16 and RL uses a batch size of 12, with both performing a gradient update once per step. We set the learning rate to $1\mathrm{e}{-6}$ for both SFT  (10 epochs) and RL  (2 epochs). We evaluate mainstream MLLMs under a unified protocol on our dataset. We use Qwen3-Coder-Plus \cite{qwen3} as the judge.

To comprehensively evaluate E-VAds, we will compare it with a range of SOTA multi-modal models, grouped into three categories: the current leading closed-source models, GPT-5.2 \cite{GPT52} and Gemini3-Flash \cite{Gemini3}; general instruction-tuned models, including Qwen3-Omni \cite{qwen3omni}, InternVL3 \cite{internvl3}, Keye-VL \cite{keye-vl}, Mimo-VL \cite{mimo}, VideoLlama3 \cite{videollamav3}, and the Qwen3-VL \cite{qwen3vl} / Qwen2.5-VL \cite{qwen25vl} series; and strong reasoning models, namely the thinking versions of Qwen3-Omni, InternVL3.5 \cite{internvl35}, Keye1.5-VL \cite{keye15}, Mimo-VL, and Qwen3-VL, which are designed to enhance logical reasoning capability. For GPT-5.2, we use 48 frames, while other MLLMs use 2 FPS.

Furthermore, to establish a human performance baseline, we randomly sampled 400 questions from the test set for manual evaluation. Two independent annotators were required to answer these questions by watching the videos only, without access to external ASR/OCR transcripts. For each question, annotators were tasked with providing both a terminal answer and the corresponding reasoning process within a strict 5-minute time limit. The final human performance, denoted as "Expert$^*$" in our results, is reported as the average score of these two annotators evaluated under the same LLM-as-a-judge protocol.

\subsection{Results and Obervations}

We comprehensively evaluate model performance on E-VAds using the strictness metrics defined in Section~\ref{sec:RL} and Qwen3-Coder-Plus as the judge, with results summarized in Table~\ref{tab:compare}. We draw the following observations. And we give a case on Appendix.~\ref{app:case_study}.

\textbf{(a) E-VAds-R1 delivers the strongest improvement among open models by explicitly training reasoning.}
E-VAds-R1 (8B) substantially improves over its base model, Qwen3-VL-8B, raising the \textbf{ALL} score from 0.153 to 0.320 under S, which is a \textbf{109.2\%} relative gain. The largest gain is on RC, where E-VAds-R1 reaches 0.279 (S), about 16$\times$ higher than the base model (0.017), and higher than GPT~5.2 (0.105) and Gemini3-Flash (0.222). This suggests that our RL training, which supervises the reasoning process, markedly improves commercial judgment.

\textbf{(b) Human experts remain a strong upper bound, highlighting the difficulty of commercial videos' reasoning.}
Table~\ref{tab:compare} shows a substantial gap between SOTA MLLMs and human experts, especially on higher-order reasoning. Human experts achieve 0.535 (S) and 0.871 (R-5) on the \textbf{ALL} metric, while the best closed-source model, Gemini3-Flash, reaches 0.350 (S). The gap is largest on Marketing Logic (ML) and Consumer Insight (CI), where even the strongest models rarely exceed 0.200 under S, indicating that current models still lack the domain expertise that is needed to infer persuasion strategies and audience psychology in e-commerce ads.

\textbf{(c) Closed-source models lead overall but still fall short of human-level commercial understanding.}
Closed-source models, including Gemini3-Flash and GPT~5.2, outperform most open-source baselines on E-VAds. Gemini3-Flash achieves the best \textbf{ALL} score at 0.350 (S) and 0.761 (R-5), and ranks first on CM with 0.391, suggesting stronger alignment between ASR narration and fast-changing visuals. GPT~5.2 is competitive but weaker than Gemini3-Flash, and both remain below human performance on commercial.

\begin{table}[t!]
\centering
\caption{Ablation study of E-VAds-R1 based on Qwen2.5-VL-7B. S,R3,R5 means the different scoring method defined in Sec.~\ref{sec:RL}}
\renewcommand{\arraystretch}{0.9}
\renewcommand{\tabcolsep}{1.2mm}
\begin{tabular}{c|c|c|c|ccc}
\toprule
\textbf{EXP} & \textbf{SFT} & \textbf{Think} & \textbf{Answer} & \textbf{S}     & \textbf{R3}    & \textbf{R5}   \\
\hline
Baseline     & –            & –              & –               & .111          & .369          & .489         \\
\hline
a1           & A            & –              & –               & .136          & .420           & .566         \\
a2           & T$\rightarrow$A         & –              & –               & .131          & .418          & .568         \\
\hline
b1           & –            & –              & S               & .136          & .393          & .511         \\
b2           & –            & –              & R3              & .141          & .458          & .559         \\
b3           & –            & –              & R5              & .152          & .479          & .614         \\
\hline
c1           & T$\rightarrow$A         & –              & R5              & .163          & .479          & .646         \\
c2           & T$\rightarrow$A         & –              & G               & .180           & .496          & {\ul .668}   \\
\hline
d1           & T$\rightarrow$A         & R5$^{0.5}$     & R5$^{0.5}$      & .169          & .486          & .658         \\
d2           & T$\rightarrow$A         & G$^{0.5}$      & G$^{0.5}$       & {\ul .192}    & {\ul .497}    & .666         \\
d3           & T$\rightarrow$A         & G$^{0.2}$      & G$^{0.8}$       & \textbf{.193} & \textbf{.501} & \textbf{.680}\\
\bottomrule
\end{tabular}
\label{tab:ablation}
\end{table}

\textbf{(d) Standard instruction-tuned open-source models show a clear reasoning bottleneck that scaling alone does not fix.}
Models such as VideoLlama3 (7B) and InternVL3 (8B) are adequate on Basic Perception (0.306 and 0.531) but collapse on ML and CI, where several score 0.000 under S. Scaling to 32B, as in Qwen2.5-VL, yields limited benefit and reaches only 0.001 on ML, suggesting that parameter scaling alone cannot solve the multi-step reasoning required in e-commerce. In contrast, thinking variants improve performance: Qwen3-VL-8B-Thinking increases the \textbf{ALL} S score from 0.153 to 0.186, indicating the importance of structured reasoning.

\subsection{Impact of different training strategies}
We conduct comprehensive ablation studies to evaluate the impact of each training component, as shown in \tabref{tab:ablation}. The Baseline is Qwen2.5-VL 7B.

\textbf{(a) SFT Strategies.} We compare two primary SFT configurations: (a1) mapping questions directly to answers (A), and (a2) a thinking-based flow (T$\rightarrow$A).
Results show only marginal performance gaps between these variants.
This suggests that the SFT stage primarily serves to align output formats and inject basic domain knowledge, whereas complex e-commerce reasoning capabilities are predominantly developed during the RL stage.

\textbf{(b, c) Reward Design.}
During the RL stage, we evaluate reward functions with different strictness levels, including $S$, $R3$, and $R5$, as well as their ensemble variant $G$. Under single-granularity reward training, a more lenient reward consistently performs better than a stricter one (b3 \textgreater b2 \textgreater b1), suggesting that overly strict rewards produce sparse supervision in dense multi-modal settings and hinder exploration. Using the multi-granularity reward $G$ further improves performance (c2 \textgreater c1), indicating that our design can better handle the complexity of e-commerce video tasks.

\textbf{(d) Thinking and Weighting.} We further encourage the model to generate explicit reasoning process and reward it with multi-modal evidence. The results provide three insights. (i)The reasoning process improves commercial understanding (d1 \textgreater c1). (ii) Our multi-grained reward yields further gains under the reasoning paradigm, as ; and it provides complementary supervision at different strictness levels, which reduces reward sparsity while still enforcing evidence grounding, matching the dense and noisy signals in E-VAds videos. (iii) Assigning a larger weight to the answer score improves performance, as stronger answer-level optimization prevents the model from producing plausible but weakly supported answers. Overall, these findings support our training and reward design choices and confirm that E-VAds is a challenging frontier in video understanding.


\begin{table}[t!]
\centering
\caption{Performance variation of models under adding ASR text.}
\renewcommand{\arraystretch}{0.9}
\renewcommand{\tabcolsep}{1.0mm}
\begin{tabular}{c|ccc|ccc}
\toprule
\textbf{Model}                    & \multicolumn{3}{c|}{\textbf{Qwen3-Omini-Thinking}}                                & \multicolumn{3}{c}{\textbf{Keye1.5-VL}}                                                 \\
\hline
\textbf{Metric}                      & \textbf{S}                  & \textbf{R3}                 & \textbf{R5}                 & \textbf{S}                  & \textbf{R3}                 & \textbf{R5}       
\\
ASR                              & .092                        & .346                        & .549                        & .092                        & .338                        & .536                        \\
Video                              & .196                        & .479                        & .659                        & .179                        & .458                        & .623                        \\
Video+ASR                          & .207                        & .498                        & .675                        & .206                        & .482                        & .650                        \\
\hline
{\color[HTML]{70AD48} $\Delta\uparrow$} & {\color[HTML]{70AD48} .060} & {\color[HTML]{70AD48} .040} & {\color[HTML]{70AD48} .024} & {\color[HTML]{70AD48} .152} & {\color[HTML]{70AD48} .052} & {\color[HTML]{70AD48} .044} \\
\hline
\textbf{Model}                    & \multicolumn{3}{c|}{\textbf{Qwen3-Vl-8B-Thinking}}                                       & \multicolumn{3}{c}{\textbf{E-VAds-R1}}                                           \\
\hline
\textbf{Metric}                      & \textbf{S}                  & \textbf{R3}                 & \textbf{R5}                 & \textbf{S}                  & \textbf{R3}                 & \textbf{R5}                 \\
ASR                              & .081                        & .328                        & .528                        & .162                        & .404                        & .597                        \\
Video                              & .160                        & .458                        & .634                        & .257                        & .532                        & .701                        \\
Video+ASR                          & .186                        & .482                        & .660                        & .320                        & .577                        & .736                        \\
\hline
{\color[HTML]{70AD48} $\Delta\uparrow$} & {\color[HTML]{70AD48} .166} & {\color[HTML]{70AD48} .054} & {\color[HTML]{70AD48} .042} & {\color[HTML]{70AD48} .245} & {\color[HTML]{70AD48} .085} & {\color[HTML]{70AD48} .050}\\
\bottomrule
\end{tabular}
\label{tab:ASR}
\end{table}

\subsection{Impact of ASR and Video inputs}
To assess the necessity of ASR transcripts and visual inputs for tasks in E-VAds, we conduct a comprehensive ablation study across four models, evaluating \textit{ASR}, \textit{Video}, and \textit{Video+ASR} configurations, as detailed in Table~\ref{tab:ASR}. Our results reveal that the visual modality remains indispensable: model performance exhibits a clear progressive improvement when transitioning from \textit{ASR} to \textit{Video}, and further to \textit{Video+ASR}. This key finding underscores the foundational role of visual information in the task. Furthermore, compared to \textit{Video} inputs, incorporating ASR transcripts consistently yields substantial performance gains across all evaluated MLLMs, with the Omni model demonstrating the smallest improvement. Collectively, these results highlight that E-VAds effectively necessitates joint audio-visual reasoning, reinforcing its value as a benchmark for evaluating multi-modal evidence-based commercial understanding.

\section{Conclusion}

In this study, we present E-VAds, a comprehensive benchmark designed to evaluate multimodal large language models on conversion-oriented e-commerce short videos. These videos are characterized by rapidly synchronized visual, auditory, and textual cues. We constructed this benchmark through a rigorous pipeline that combines multi-agent automated generation with expert human verification. The final dataset contains 3,961 carefully curated videos that span diverse product categories. Accompanying these videos are 19,785 open-ended question-answer pairs. These evaluation items are systematically organized into five distinct task categories. The categories include Basic Perception, Cross-Modal Detection, Marketing Logic, Consumer Insight, and Regulatory Compliance. We quantified the informational complexity of this domain through a dedicated multi-modal density assessment framework. The quantitative results demonstrate that E-VAds exhibits substantially higher visual dynamic density, audio speech intensity, and textual overlay frequency when compared to mainstream video understanding datasets. This empirical evidence establishes e-commerce video question answering as a uniquely challenging research frontier. The domain requires models to process tightly synchronized signals and execute multi-step commercial reasoning simultaneously. To address the supervision sparsity inherent in these open-ended commercial reasoning tasks, we developed E-VAds-R1. This reinforcement learning framework incorporates a multi-grained reward mechanism that strictly grounds model outputs in multi-modal evidence. The proposed MG-GRPO strategy delivers smooth optimization signals during early training phases. It also establishes a non-linear incentive structure that drives the model toward expert-level precision. Experimental evaluations confirm that E-VAds-R1 achieves remarkable performance improvements while utilizing only a few hundred training samples.

\section*{Impact Statement}
\textbf{Benchmark and research impact.} E-VAds provides a challenging, standardized benchmark for conversion-oriented e-commerce short videos under modality-dense signals, facilitating evaluation of MLLMs on fine-grained perception, cross-modal grounding, and commercial-intent reasoning.

\textbf{Practical impact.} The dataset and evaluation can support applications such as e-commerce video understanding, ad content analysis, retrieval and summarization, and assistance tools that help users better access key product information.

\textbf{Limitations and inherent video bias.} Our construction pipeline employs comprehensive bias mitigation strategies that encompass a multi-agent adversarial QA generation framework, strict cross-modal evidence traceability constraints, and rigorous human expert review. These measures effectively reduce biases that emerge during the annotation process. We acknowledge that inherent biases within the source e-commerce videos remain difficult to eliminate completely. Such characteristics encompass commercial promotion orientation, cultural and regional preferences, distinct creator presentation styles, and platform specific product selection patterns. These characteristics originate from the real-world nature of the commercial video content. Complete debiasing at the raw video level therefore remains an open technical challenge. We explicitly document this limitation to inform future research efforts that aim to develop more balanced and representative commercial benchmarks.

\textbf{Acknowledgement.}
This work was supported by alibaba Group through Alibaba Research Intern Program.

\bibliography{example_paper}
\bibliographystyle{icml2026}

\newpage
\appendix
\onecolumn

\section{Distribution of datasets}
\label{app:distribution}
\begin{figure*}[t!]
  \centering
  \includegraphics[width=0.85\textwidth]{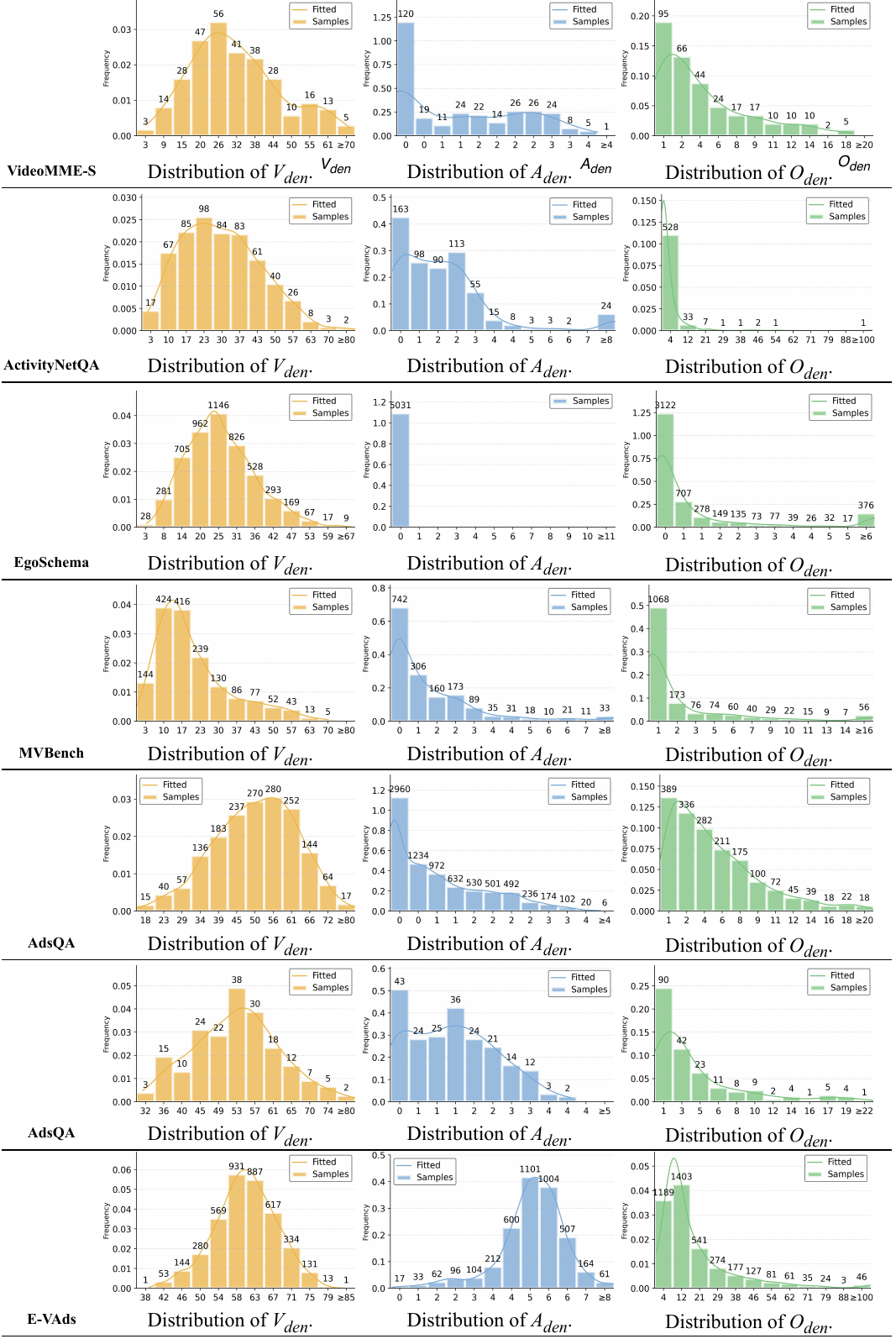}
  \caption{Detailed distributions of multi-modal information density metrics ($V_{den}$, $A_{den}$, and $O_{den}$) across datasets.}
  \label{fig:distribution}
\end{figure*}

We visualize the multi-modal information density distributions of E-VAds against VideoMME-S \cite{videomme}, ActivityNetQA \cite{activitynetqa}, EgoSchema \cite{egoschema}, MVBench \cite{mvbench}, VideoAds \cite{videoads}, and AdsQA \cite{adsqa}.
E-VAds exhibits consistently higher multimodal information density as shown in Fig.~\ref{fig:distribution} and Tab.~\ref{app:benchmark_comparison}.
For $V_{den}$, its distribution is clearly shifted toward larger values than those of general-domain videos such as ActivityNetQA, which reflects faster editing and more frequent shot changes in e-commerce short videos.
For $A_{den}$, E-VAds concentrates on the high-density region in terms of ASR word frequency, whereas the compared benchmarks are flatter or biased toward low-frequency regions, indicating that e-commerce videos contain substantially denser spoken content.
For $O_{den}$, E-VAds shows the most pronounced advantage in OCR density because e-commerce videos frequently overlay stylized captions, key selling points, and price tags, which results in markedly more text per frame than non-advertising datasets.
\begin{table*}[h]
\small
    \centering
    \caption{Comparison between our proposed E-VAds and other existing video benchmarks.}
    \renewcommand{\arraystretch}{1.2}
    \renewcommand{\tabcolsep}{5.0mm}
    \begin{tabular}{c|cccc}
        \toprule
\textbf{Benchmark} & \textbf{Nums} & \textbf{Avg dur (S)} & \textbf{Total dur (H)} & \textbf{Avg size (MB)} \\
\hline
VideoMME-short     & 300           & 80.9                 & 6.64                   & 9.87                   \\
MVBench            & 1,629          & 23.94                & 10.83                  & 6.19                   \\
ActivityNetQA      & 574           & 107.8                & 17.06                  & 23.77                  \\
EgoSchema          & 5,031          & 179.76               & 251.55                 & 20.08                  \\
AdsQA              & 1,704          & 51.89                & 24.56                  & 17.6                   \\
VideoAds           & 200           & 79.67                & 4.26                   & 7.41                   \\
\hline
E-VAds  & 3,961          & 45.48                & 50.04                  & 9.99                  \\
        \bottomrule
    \end{tabular}
    \label{app:benchmark_comparison}
\end{table*}

\section{Task category prompts}
\label{appendix:task_prompts}
\begin{figure*}[t!]
  \centering
  \includegraphics[width=0.96\textwidth]{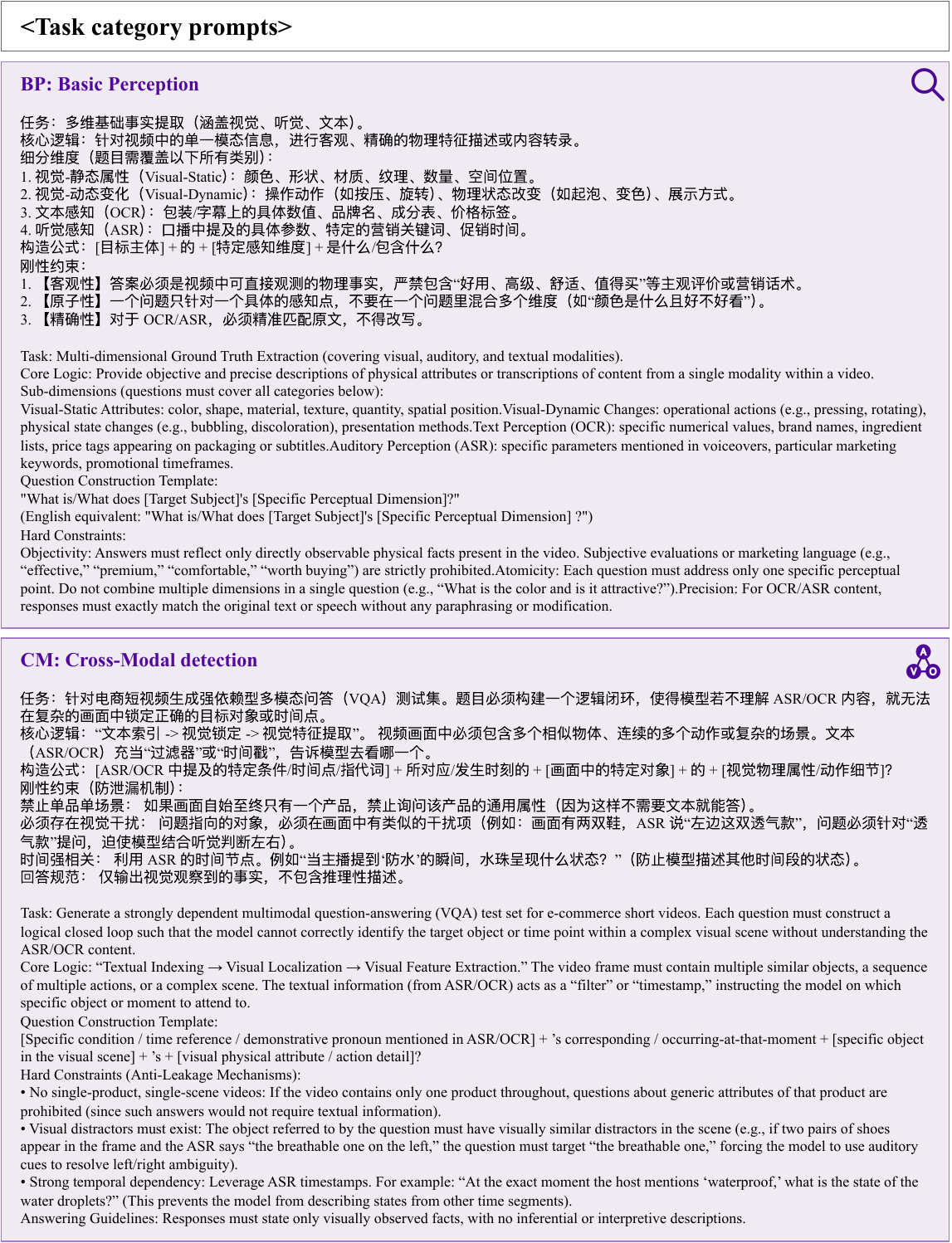}
  \caption{Task definitions and prompt instructions for Basic Perception (BP) and Cross-modal Matching (CM).}
  \label{fig:task_1}
\end{figure*}

\begin{figure*}[t!]
  \centering
  \includegraphics[width=0.96\textwidth]{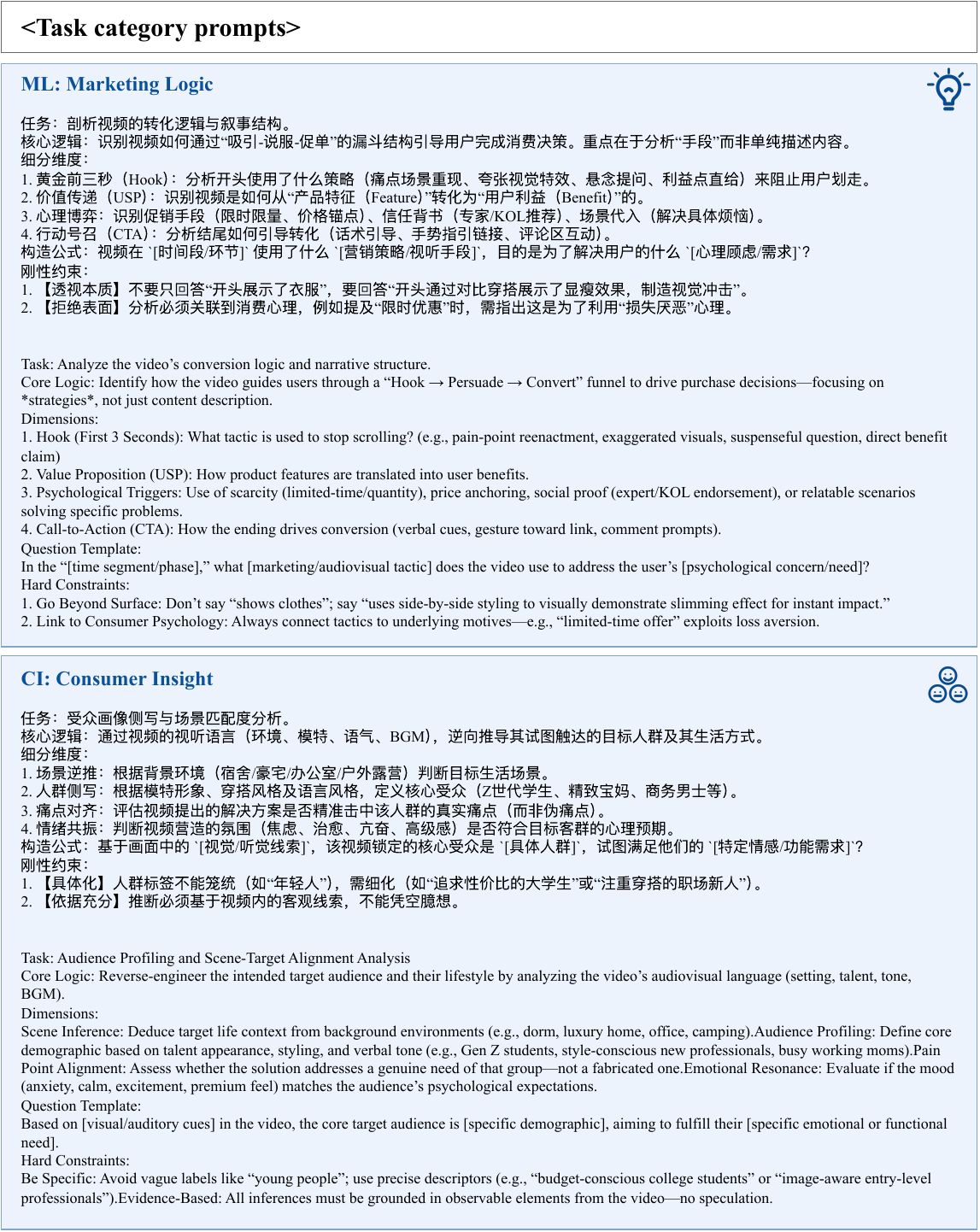}
  \caption{Task definitions and prompt instructions for Marketing Logic (ML) and Consumer Insight (CI).}
  \label{fig:task_2}
\end{figure*}

\begin{figure*}[t!]
  \centering
  \includegraphics[width=0.96\textwidth]{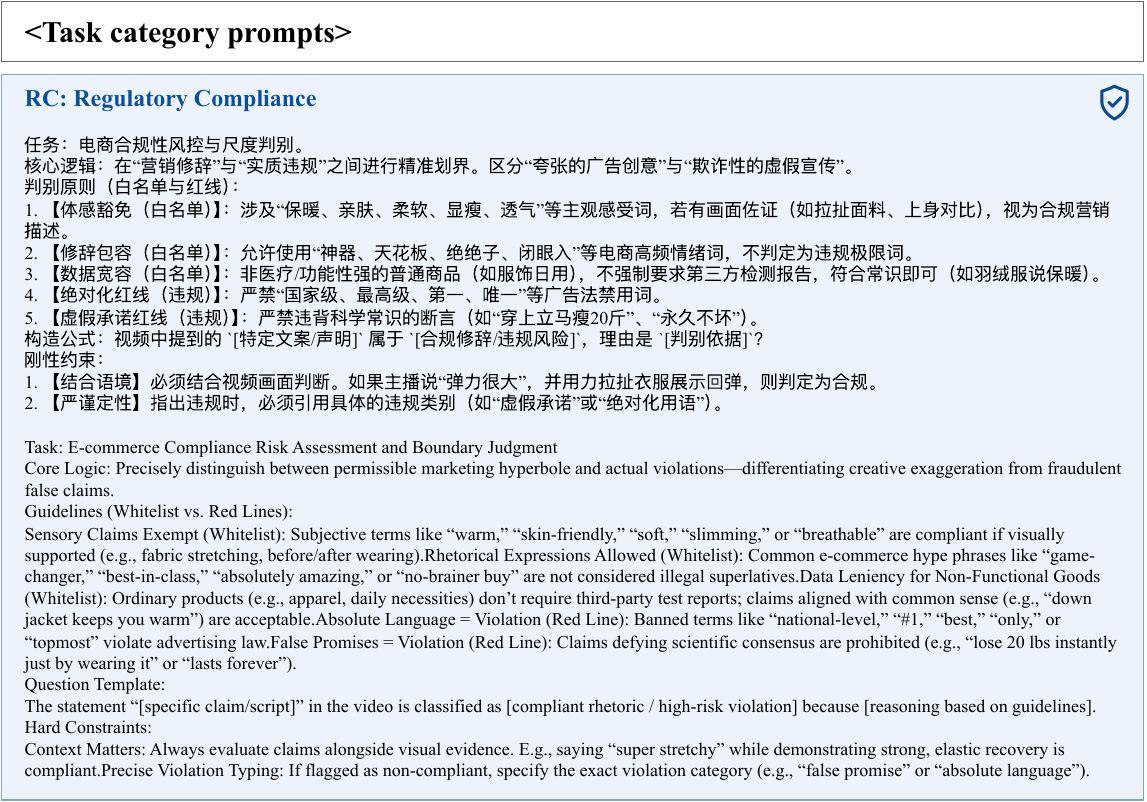}
  \caption{Task definitions and prompt instructions for Regulatory Compliance (RC).}
  \label{fig:task_3}
\end{figure*}

This section presents the core prompt logic that we use for automated annotation across five tasks as shown in Fig.~\ref{fig:task_1},~\ref{fig:task_2} and ~\ref{fig:task_3}.
For Basic Perception (BP), the prompt enforces objectivity by requiring the model to extract only physical attributes such as color, material, and numeric values, while explicitly forbidding subjective judgments.
For Cross-modal Matching (CM), the prompt compels the model to ground references from ASR and OCR, such as deictic mentions like ``this one'', to a specific visual entity.
For Marketing Logic (ML), the prompt follows a funnel-oriented analysis that guides the model to identify early hooks within the first few seconds, map USPs to user benefits, and recognize the design of calls to action.
For Consumer Insight (CI), the prompt encourages backward inference by using cues from the scene context, presenter style, and background music to derive a concrete target-audience profile.
For Regulatory Compliance (RC), the prompt implements a red-line and whitelist scheme that distinguishes permissible promotional rhetoric from illegal absolute claims, for example separating ``miracle product'' from terms such as ``national-level'' or ``No.1''.

\section{Details about multi-agent annotation system}
\label{appendix:multiagent}
\begin{figure*}[t!]
  \centering
  \includegraphics[width=0.96\textwidth]{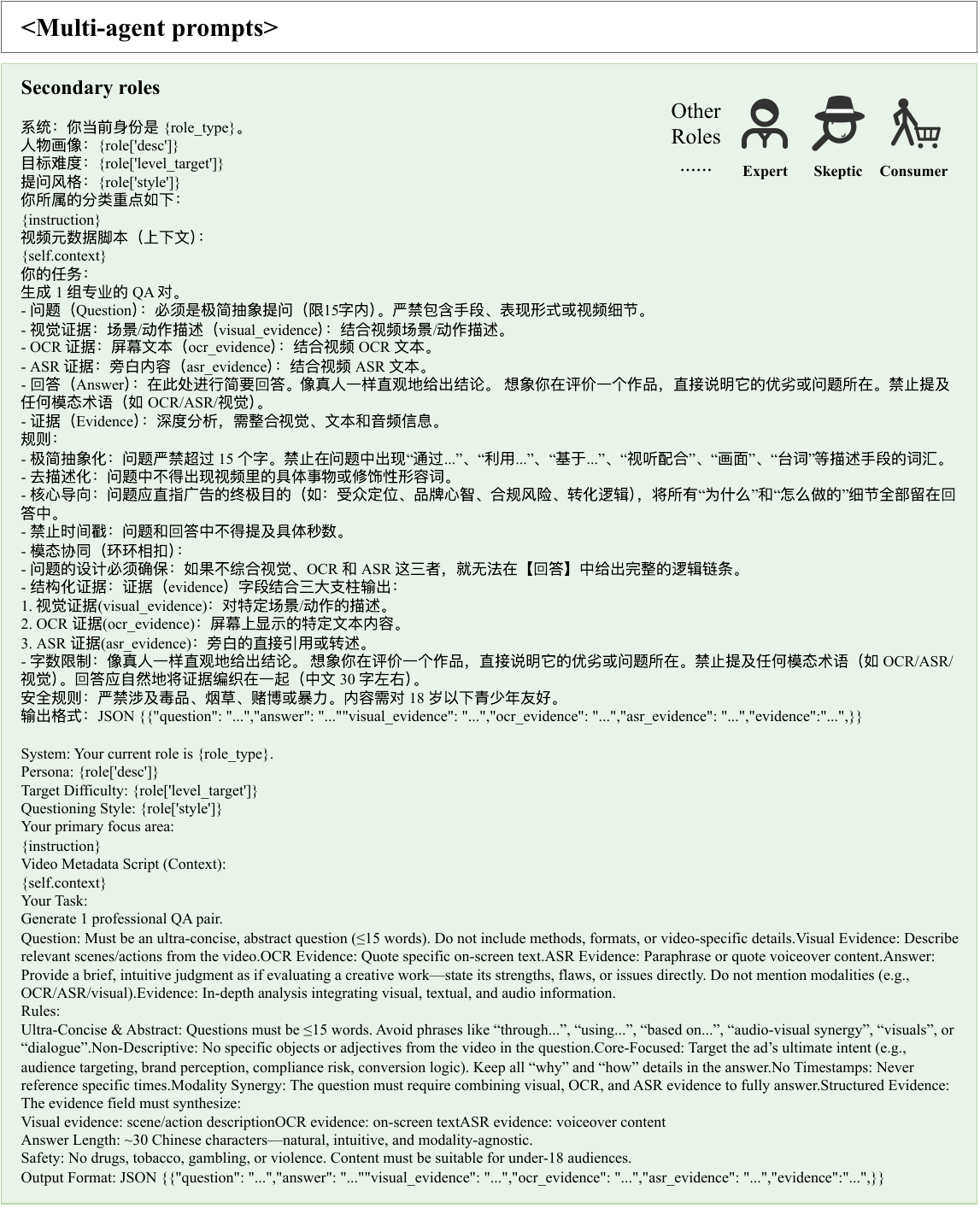}
  \caption{Prompt structure for secondary-role agents in the multi-agent system.}
  \label{fig:sec-role}
\end{figure*}

\begin{figure*}[t!]
  \centering
  \includegraphics[width=0.96\textwidth]{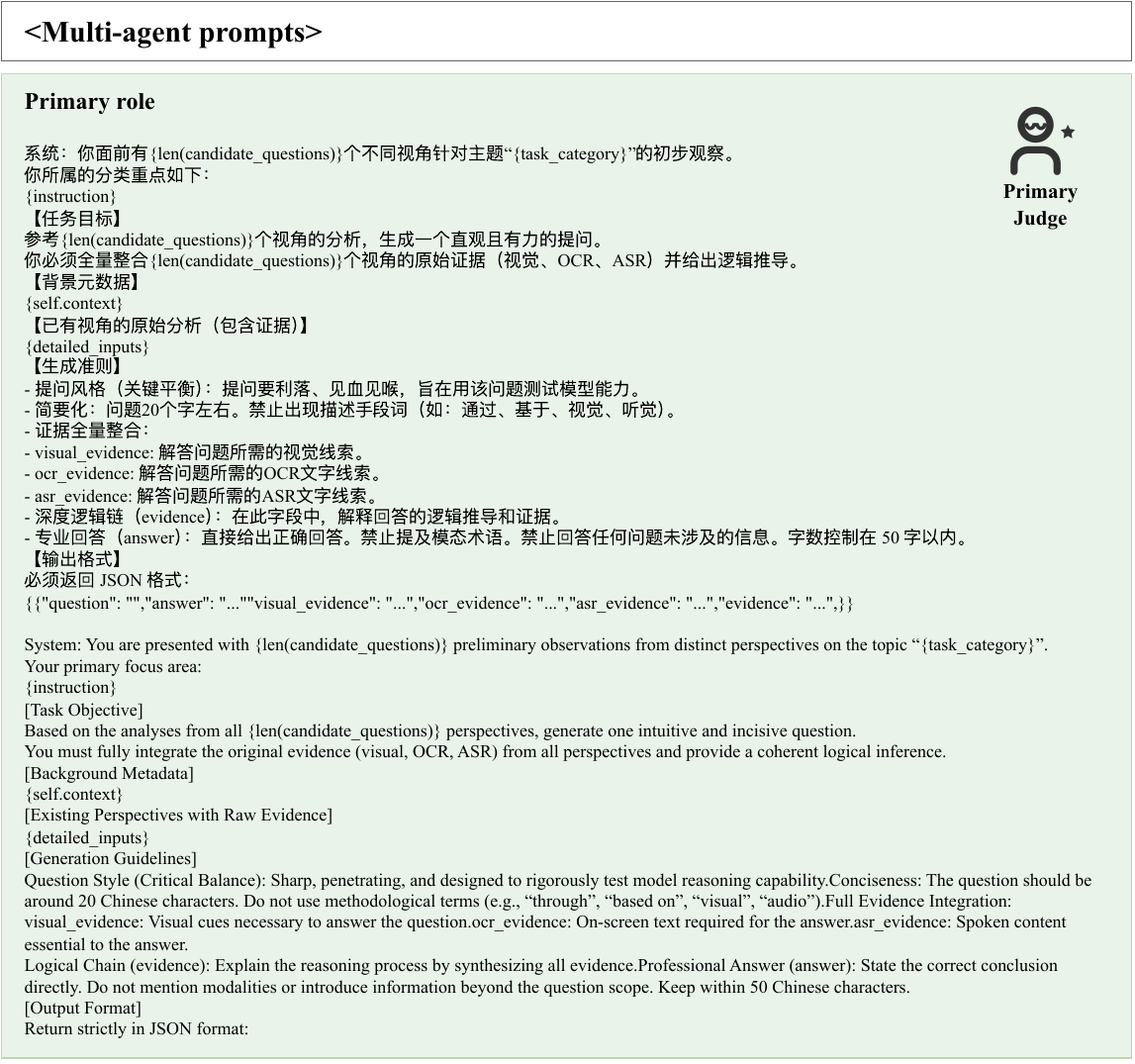}
  \caption{Integrated prompt for the primary judge agent in the multi-agent system.}
  \label{fig:pri-role}
\end{figure*}

We use a multi-agent collaboration framework to improve the quality and diversity of the generated QA pairs (Fig.~\ref{fig:sec-role} and ~\ref{fig:pri-role}).
For secondary roles. We instantiate role-specific agents, where each agent asks questions that match its persona.
We impose strict constraints such that each question is limited to 15 words and must not contain descriptive terms, which encourages abstract and challenging queries.
Next, a primary judge agent aggregates raw evidence from all modalities, including visual content, OCR, and ASR.
It consolidates multi-perspective observations into a single focused question and produces an explicit evidence chain that links multimodal cues to the final answer.
In practice, to reduce annotation cost, we use strong Closed-source models. Secondary-role agents are instantiated with Gemini~3 Flash, while the primary judge is instantiated with Gemini~3 Pro.
\paragraph{Multi-agent QA generation.}
We generate QA pairs with a multi-agent collaborative annotation system. Each agent adopts a distinct commercial persona to simulate real business viewpoints and to convert the time-aligned evidence chain into challenging, evidence-grounded questions. Annotations are organized as a virtual round table led by a \textbf{primary judge} and supported by multiple secondary roles.

For perception-oriented tasks, annotators extract cues from five complementary perspectives with difficulty levels from \textit{L1} to \textit{L3}:
\begin{itemize}[noitemsep, topsep=2pt, leftmargin=*]
    \item \textbf{Physical attributes}: color, material, shape, size, quantity, and motion.
    \item \textbf{Symbolic information}: brand, model, numbers, price, keywords, and contact information.
    \item \textbf{Relational evidence}: text--object grounding and vision--narration alignment.
    \item \textbf{Environmental context}: scene, location, weather, and landmarks.
    \item \textbf{Actionable behaviors}: operations such as unboxing, applying, pressing, stretching, step-by-step demonstrations, and user feedback (posture and facial expressions).
\end{itemize}

For reasoning tasks, we use personas with increasing difficulty from \textit{L1} to \textit{L5} (Higher levels indicate greater reasoning complexity):
\begin{enumerate}[noitemsep, topsep=2pt, leftmargin=*]
    \item \textbf{Consumer} (\textit{L1--L3}): focuses on observable experience and perceived authenticity.
    \item \textbf{Pragmatist} (\textit{L2--L3}): emphasizes functionality, value, usage steps, price anchoring, and pain point matching.
    \item \textbf{Skeptic} (\textit{L2--L4}): checks for inconsistencies or missing information across vision, ASR, and OCR.
    \item \textbf{Expert} (\textit{L3--L5}): decomposes persuasive logic and marketing positioning.
    \item \textbf{Creative Director} (\textit{L4--L5}): analyzes audiovisual language and narrative structure.
\end{enumerate}

\section{Manual Check and Quality Control}
\label{appendix:manual_check}
\begin{figure*}[t!]
  \centering
  \includegraphics[width=0.96\textwidth]{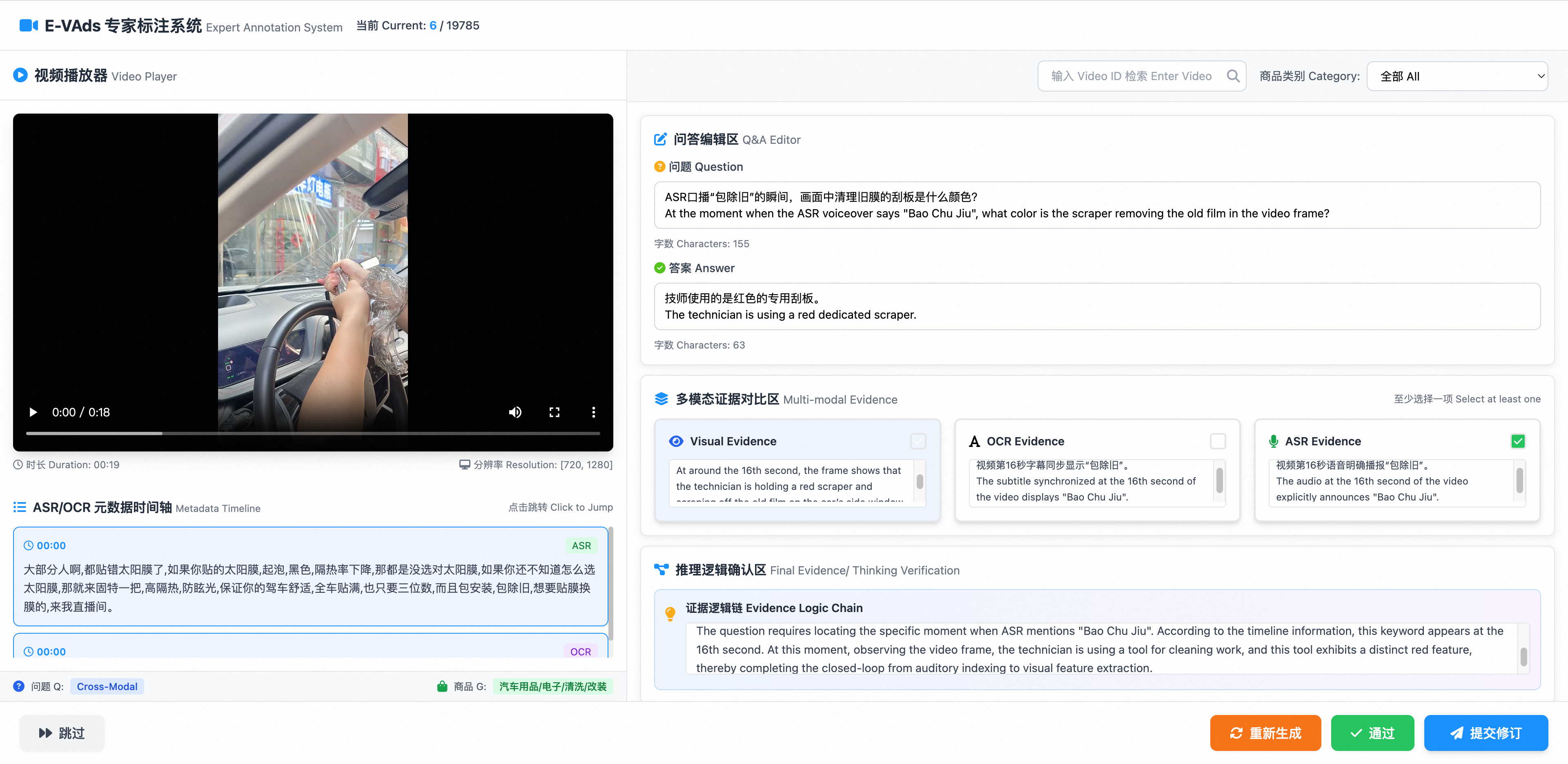}
  \caption{E-VAds Annotation System.}
  \label{fig:system}
\end{figure*}

To ensure the rigor, precision, and commercial relevance of the E-VAds benchmark, we implement a multi-stage human verification pipeline as shown in Fig.~\ref{fig:system}. The process involves five professionally trained annotators and two senior researchers who serve as lead auditors to resolve disputes and perform final quality spot-checks.

Annotators evaluate each sample based on four fundamental pillars. First, \textit{Accuracy} requires that answers remain strictly faithful to video facts and ASR/OCR references. Second, \textit{Traceability}, which serves as the core principle, mandates that all responses be derived exclusively from provided multimodal evidence to prevent hallucinations based on external common sense. Third, \textit{Discriminability} ensures that questions are sufficiently challenging such that they cannot be solved via shortcuts or linguistic biases without viewing the video. Finally, \textit{Commercial Relevance} demands that answers reflect professional e-commerce insights, such as identifying a ``price anchoring strategy'' rather than merely describing a discount.

The verification workflow for the five core tasks (BP, CM, ML, CI, and RC) follows a recursive refinement logic. 
During the initial review, annotators assess the QA pairs and their corresponding evidence chains. 
While valid entries are immediately accepted, any item exhibiting logical inconsistencies or misaligned evidence triggers a multi-agent regeneration process. 
We enforce a cycle-based elimination mechanism that allows for a maximum of three regeneration attempts. If a task remains unsatisfactory after the third iteration, it undergoes manual correction by human experts to ensure the ultimate quality and integrity of the dataset.

This rigorous validation process ultimately produced approximately 19,785 high-quality QA pairs. The systematic approach effectively ensures the robustness, interpretability, and professional depth of each entry in E-VAds.

\section{LLM as Judge Prompt}
\label{appendix:judge_prompt}
Traditional lexical metrics such as BLEU \cite{bleu} and ROUGE \cite{rouge} exhibit significant limitations in the context of e-commerce video understanding because they prioritize surface-level word overlap rather than semantic accuracy or business logic. These metrics often fail to distinguish between valid paraphrasing and factual hallucinations that arise from misinterpreting OCR or ASR metadata. 

To address this, we introduce an LLM-as-a-judge \cite{llmasjudge} mechanism that simulates the perspective of a professional e-commerce analyst. This framework performs deep semantic verification by cross-referencing model outputs with video metadata, clues and ground-truth answers. By focusing on whether the provided evidence aligns with the actual video content, the judge effectively identifies logical disconnects that traditional metrics might overlook. 

\begin{figure*}[t!]
  \centering
  \includegraphics[width=0.96\textwidth]{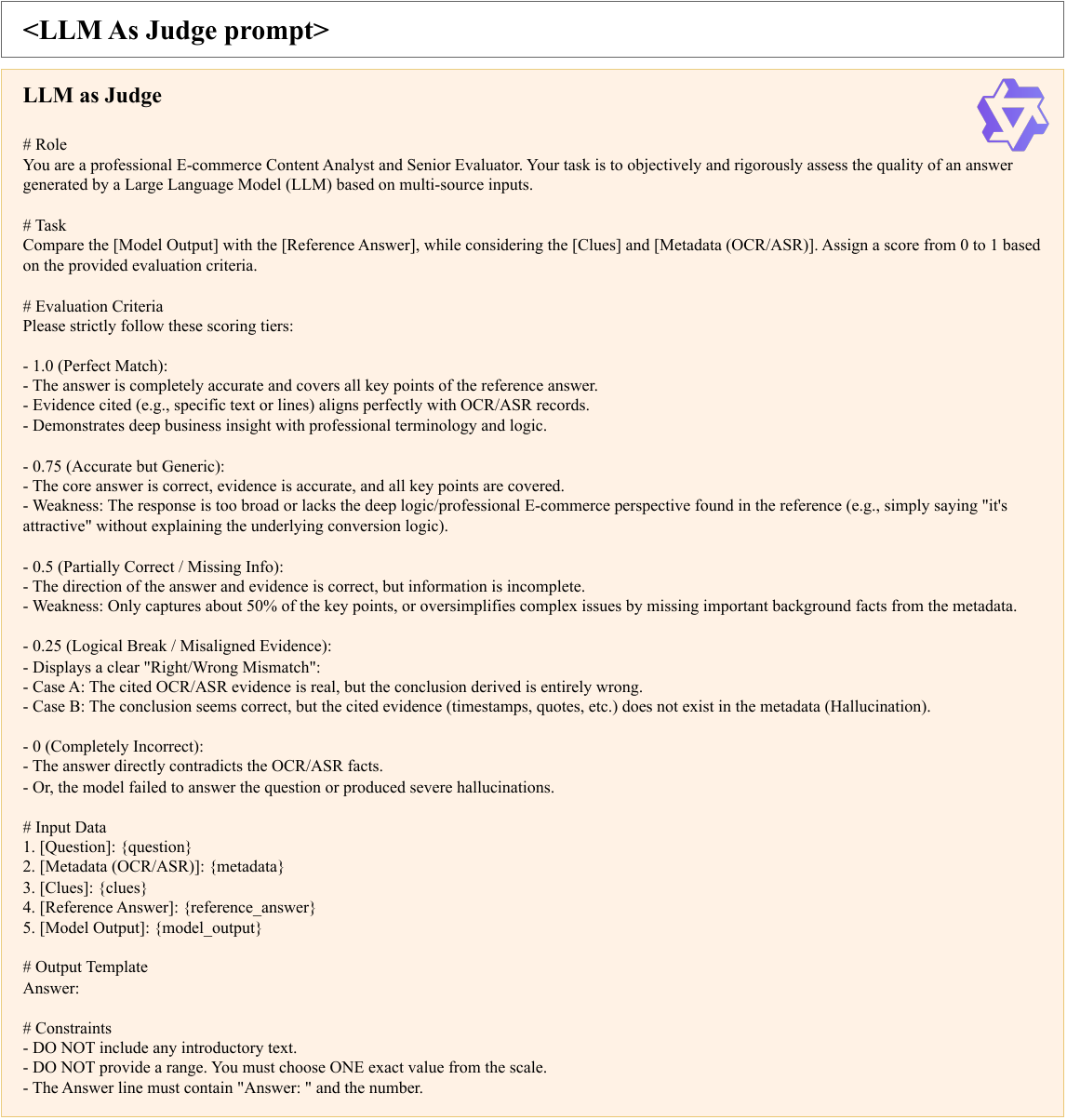}
  \caption{Evaluation prompt and scoring rubric for LLM-as-a-Judge.}
  \label{fig:llmasjudge}
\end{figure*}

As illustrated in Fig.~\ref{fig:llmasjudge}, the evaluator assigns a score from 0 to 1 based on five granular tiers: 
\begin{itemize}
    \item \textbf{1.0 (Perfect Match):} The response is accurate and professional with evidence that perfectly aligns with metadata.
    \item \textbf{0.75 (Accurate but Generic):} Core insights are correct but lack the depth of professional business analysis.
    \item \textbf{0.5 (artially Correct / Missing Info):} Only approximately half of the key points are captured or major background facts are omitted.
    \item \textbf{0.25 (Logical Break / Misaligned Evidence):} The conclusion appears plausible but is supported by incorrect evidence which constitutes a factual hallucination.
    \item \textbf{0 (Completely Incorrect):} The response entirely deviates from the facts or fails to follow instructions.
\end{itemize}
This multi-tiered approach provides a more precise and objective assessment of MLLM performance in high-density information environments.

\section{Answer Prompts}
\label{appendix:sft_rl_prompts}
\begin{figure*}[t!]
  \centering
  \includegraphics[width=0.96\textwidth]{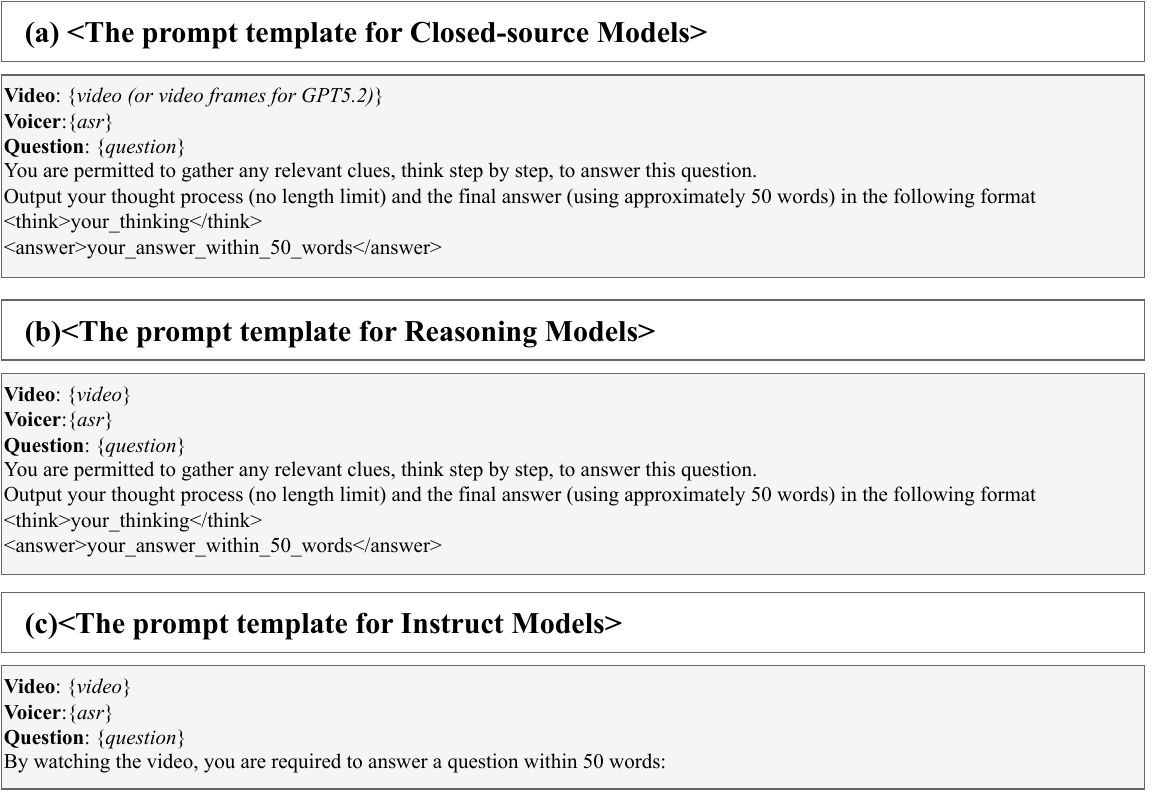}
  \caption{Prompt Format for Inference.}
  \label{fig:Prompt}
\end{figure*}

For close-source models, we use this prompt as shown in Fig.~\ref{fig:Prompt} a.
For reasoning models, we use this prompt as shown in Fig.~\ref{fig:Prompt} b.
For instruct models, we use this prompt as shown in Fig.~\ref{fig:Prompt} c.

\section{Case Study}
\label{app:case_study}
\begin{figure*}[t!]
  \centering
  \includegraphics[width=0.96\textwidth]{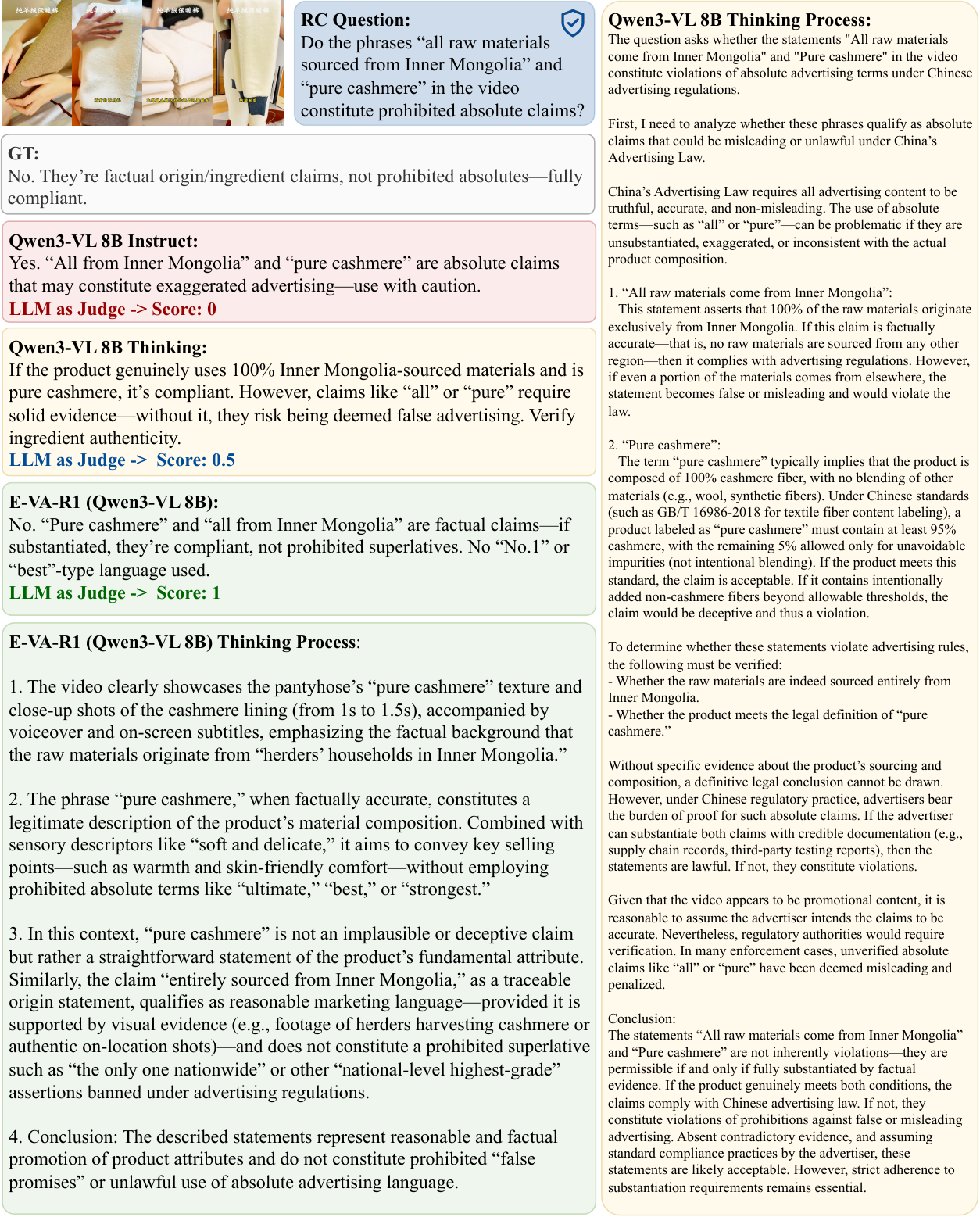}
  \caption{Case Study.}
  \label{fig:case}
\end{figure*}

As show in Fig.~\ref{fig:case}, while general MLLMs often fail by being overly cautious or indecisive regarding regulatory nuances, E-VAds-R1 demonstrates expert-level judgment. It correctly distinguishes factual ingredient/origin claims from prohibited superlatives (e.g., "No.1" or "Best"), showing that our RL-based reasoning significantly bridges the gap between general perception and specialized commercial logic.

\end{document}